\pdfoutput=1
\documentclass[11pt]{article}
\usepackage{amsmath}
\usepackage[final]{acl}
\usepackage{times}
\usepackage{latexsym}
\usepackage[T1]{fontenc}
\usepackage[utf8]{inputenc}
\usepackage{microtype}
\usepackage{inconsolata}
\usepackage{graphicx}
\usepackage{booktabs}
\usepackage{multirow}
\usepackage{subfigure}
\usepackage{amsmath}
\usepackage{amssymb}
\usepackage{hyperref}
\usepackage{float}
\usepackage{multirow}
\usepackage{booktabs}
\usepackage{amsmath}
\title{Constraining Sequential Model Editing with Editing Anchor Compression}

\author{
Hao-Xiang Xu$^1$\thanks{Equal contribution.}, Jun-Yu Ma$^1$\footnotemark[1], Zhen-Hua Ling$^1$\thanks{Corresponding author.}, Ningyu Zhang$^2$, Jia-Chen Gu$^3$  \\
  $^1$National Engineering Research Center of Speech and Language Information Processing, \\
      University of Science and Technology of China, Hefei, China \\
  $^2$Zhejiang University \\
  $^3$University of California, Los Angeles \\
{\tt \{nh2001620,mjy1999\}@mail.ustc.edu.cn}, {\tt zhling@ustc.edu.cn}, \\ 
{\tt zhangningyu@zju.edu.cn}, {\tt gujc@ucla.edu} 
}

\begin{document}
\maketitle
\begin{abstract}
Large language models (LLMs) struggle with hallucinations due to false or outdated knowledge. Given the high resource demands of retraining these models, there is an increasing focus on developing \emph{model editing}. 
However, the general abilities of LLMs across downstream tasks are prone to significant degradation during sequential editing.
This paper statistically observes that the parameter matrix after editing exhibits a significant deviation compared to its previous state as the number of edits increases.
This serious deviation affects the original knowledge associations within LLMs and leads to the degradation of their general abilities.
To this end, a framework termed \textbf{E}diting \textbf{A}nchor \textbf{C}ompression (EAC) is proposed to constrain the deviation of the parameter matrix during sequential editing. 
It compresses the editing information by selecting editing anchors that are important in encoding new relations without deviating too much from the original matrix, thereby preserving the general abilities. 
Experiments of applying EAC to two popular editing methods on three LLMs across four tasks are conducted. 
Evaluation results show that EAC effectively minimizes unreasonable deviations caused by model editing, preserving over 70\% of the general abilities while better retaining the editing knowledge compared to the original counterpart methods\footnote{Code is available: \href{https://github.com/famoustourist/EAC}{https://github.com/famoustourist/EAC}}.
\end{abstract}

\section{Introduction}

Despite the remarkable capabilities of large language models (LLMs)~\cite{DBLP:conf/emnlp/QinZ0CYY23,DBLP:journals/corr/abs-2307-09288}, they often inevitably exhibit hallucinations due to incorrect or outdated knowledge embedded in their parameters~\cite{DBLP:journals/corr/abs-2309-01219, DBLP:journals/corr/abs-2302-12813, DBLP:journals/csur/JiLFYSXIBMF23}.
Given the significant time and expense required to retrain LLMs, there has been growing interest in \emph{model editing} (a.k.a., \emph{knowledge editing})~\cite{DBLP:conf/iclr/SinitsinPPPB20, DBLP:journals/corr/abs-2012-00363, DBLP:conf/acl/DaiDHSCW22, DBLP:conf/icml/MitchellLBMF22, DBLP:conf/nips/MengBAB22, DBLP:conf/iclr/MengSABB23, DBLP:conf/emnlp/YaoWT0LDC023, DBLP:conf/emnlp/ZhongWMPC23, DBLP:conf/icml/MaL0G24, DBLP:journals/corr/abs-2401-04700}, 
which aims to update the knowledge of LLMs cost-effectively.
Some existing methods of model editing achieve this by modifying model parameters, which can be generally divided into two categories~\cite{DBLP:journals/corr/abs-2308-07269, DBLP:conf/emnlp/YaoWT0LDC023}.
Specifically, one type is based on \emph{Meta-Learning}~\cite{DBLP:conf/emnlp/CaoAT21, DBLP:conf/acl/DaiDHSCW22}, while the other is based on \emph{Locate-then-Edit}~\cite{DBLP:conf/acl/DaiDHSCW22, DBLP:conf/nips/MengBAB22, DBLP:conf/iclr/MengSABB23}. This paper primarily focuses on the latter.

\begin{figure}[t]
  \centering
  \includegraphics[width=0.48\textwidth]{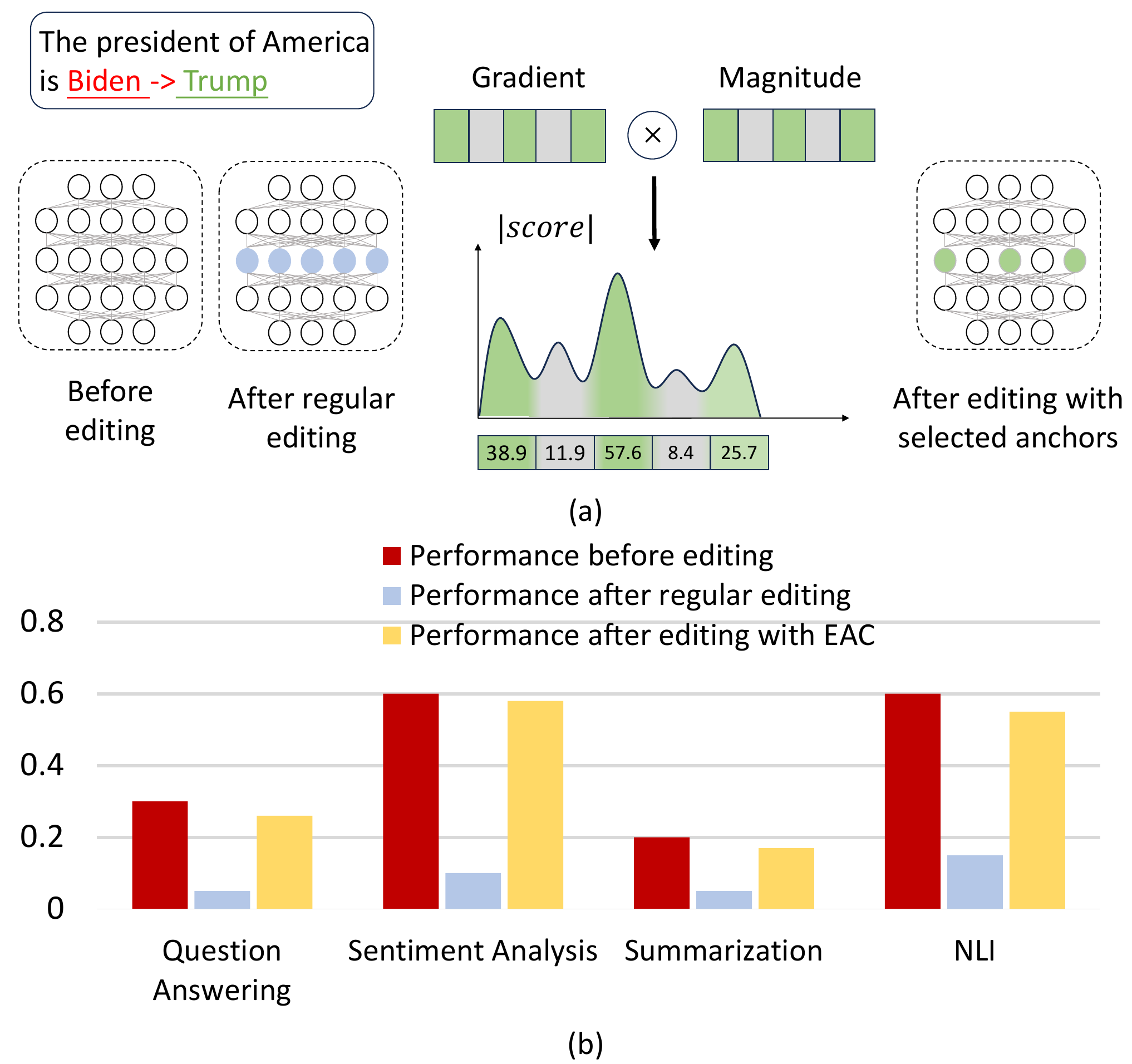}
  \vspace{-4mm}
  \caption{(a) Comparison of regular model editing and EAC. EAC compresses the editing information into the dimensions where the editing anchors are located. Here, we utilize the gradients generated during training and the magnitude of the updated knowledge vector to identify anchors. (b) Comparison of general downstream task performance before editing, after regular editing, and after constrained editing by EAC.}
  \vspace{-3mm}
  \label{demo}
\end{figure}

\emph{Sequential} model editing~\cite{DBLP:conf/emnlp/YaoWT0LDC023} can expedite the continual learning of LLMs where a series of consecutive edits are conducted.
This is very important in real-world scenarios because new knowledge continually appears, requiring the model to retain previous knowledge while conducting new edits. 
Some studies have experimentally revealed that in sequential editing, existing methods lead to a decrease in the general abilities of the model across downstream tasks~\cite{DBLP:journals/corr/abs-2401-04700, DBLP:conf/acl/GuptaRA24, DBLP:conf/acl/Yang0MLYC24, DBLP:conf/acl/HuC00024}. 
Besides, \citet{ma2024perturbation} have performed a theoretical analysis to elucidate the bottleneck of the general abilities during sequential editing.
However, previous work has not introduced an effective method that maintains editing performance while preserving general abilities in sequential editing.
This impacts model scalability and presents major challenges for continuous learning in LLMs.

In this paper, a statistical analysis is first conducted to help understand how the model is affected during sequential editing using two popular editing methods, including ROME~\cite{DBLP:conf/nips/MengBAB22} and MEMIT~\cite{DBLP:conf/iclr/MengSABB23}.
Matrix norms, particularly the L1 norm, have been shown to be effective indicators of matrix properties such as sparsity, stability, and conditioning, as evidenced by several theoretical works~\cite{kahan2013tutorial}. In our analysis of matrix norms, we observe significant deviations in the parameter matrix after sequential editing.
Besides, the semantic differences between the facts before and after editing are also visualized, and we find that the differences become larger as the deviation of the parameter matrix after editing increases.
Therefore, we assume that each edit during sequential editing not only updates the editing fact as expected but also unintentionally introduces non-trivial noise that can cause the edited model to deviate from its original semantics space.
Furthermore, the accumulation of non-trivial noise can amplify the negative impact on the general abilities of LLMs.

Inspired by these findings, a framework termed \textbf{E}diting \textbf{A}nchor \textbf{C}ompression (EAC) is proposed to constrain the deviation of the parameter matrix during sequential editing by reducing the norm of the update matrix at each step. 
As shown in Figure~\ref{demo}, EAC first selects a subset of dimension with a high product of gradient and magnitude values, namely editing anchors, that are considered crucial for encoding the new relation through a weighted gradient saliency map.
Retraining is then performed on the dimensions where these important editing anchors are located, effectively compressing the editing information.
By compressing information only in certain dimensions and leaving other dimensions unmodified, the deviation of the parameter matrix after editing is constrained. 
To further regulate changes in the L1 norm of the edited matrix to constrain the deviation, we incorporate a scored elastic net ~\cite{zou2005regularization} into the retraining process, optimizing the previously selected editing anchors.

To validate the effectiveness of the proposed EAC, experiments of applying EAC to \textbf{two popular editing methods} including ROME and MEMIT are conducted.
In addition, \textbf{three LLMs of varying sizes} including GPT2-XL~\cite{radford2019language}, LLaMA-3 (8B)~\cite{llama3} and LLaMA-2 (13B)~\cite{DBLP:journals/corr/abs-2307-09288} and \textbf{four representative tasks} including 
natural language inference~\cite{DBLP:conf/mlcw/DaganGM05}, 
summarization~\cite{gliwa-etal-2019-samsum},
open-domain question-answering~\cite{DBLP:journals/tacl/KwiatkowskiPRCP19},  
and sentiment analysis~\cite{DBLP:conf/emnlp/SocherPWCMNP13} are selected to extensively demonstrate the impact of model editing on the general abilities of LLMs. 
Experimental results demonstrate that in sequential editing, EAC can effectively preserve over 70\% of the general abilities of the model across downstream tasks and better retain the edited knowledge.

In summary, our contributions to this paper are three-fold:
(1) This paper statistically elucidates how deviations in the parameter matrix after editing are responsible for the decreased general abilities of the model across downstream tasks after sequential editing.
(2) A framework termed EAC is proposed, which ultimately aims to constrain the deviation of the parameter matrix after editing by compressing the editing information into editing anchors. 
(3) It is discovered that on models like GPT2-XL and LLaMA-3 (8B), EAC significantly preserves over 70\% of the general abilities across downstream tasks and retains the edited knowledge better.
\section{Related Work}

\paragraph{Model Editing}
Many model editing methods focus on updating knowledge in LLMs by modifying parameters, including meta-learning and locate-then-edit approaches~\cite{DBLP:conf/emnlp/YaoWT0LDC023,DBLP:journals/corr/abs-2308-07269}. Meta-learning methods train a hypernetwork to apply gradient-based updates to model parameters~\cite{DBLP:conf/emnlp/CaoAT21,DBLP:conf/iclr/MitchellLBFM22}, as in Knowledge Editor (KE) and MEND. Locate-then-edit methods identify MLPs storing factual knowledge and edit them by injecting new key-value pairs~\cite{DBLP:conf/nips/MengBAB22,DBLP:conf/iclr/MengSABB23}, leveraging the observation that these layers can act as key-value memories~\cite{DBLP:conf/emnlp/GevaSBL21, DBLP:conf/emnlp/GevaCWG22}. Additionally,~\citet{DBLP:journals/corr/abs-2012-00363} proposed constrained fine-tuning for modified facts. Recent works extend these approaches to domains such as model personality changes~\cite{DBLP:journals/corr/abs-2310-02168}, multimodal models~\cite{DBLP:conf/emnlp/0008TL0WC023}, and user privacy protection~\cite{DBLP:conf/emnlp/WuLXDW0X23}. 
Studies also focus on sequential editing scenarios, showing that as edits increase, general abilities tend to degrade~\cite{DBLP:journals/corr/abs-2401-04700,DBLP:conf/acl/0003BLZZW0H24,jiang2024neuron,fang2024alphaedit}.
Additionally, \citet{ma2024perturbation} made a theoretical analysis to elucidate that the bottleneck of the general abilities during sequential editing lies in the escalating value of the condition number of the edited matrix.

\paragraph{Saliency Analyses}
In Computer Vision (CV), extensive research on input saliency maps has contributed to explainable machine learning. Methods include pixel-space sensitivity maps~\cite{DBLP:journals/corr/SmilkovTKVW17} and class-discriminative localization~\cite{DBLP:journals/ijcv/SelvarajuCDVPB20}. Data-level saliency, also called data attribution, has been widely studied for model explanation~\cite{DBLP:journals/corr/abs-2308-03296}, efficient training~\cite{DBLP:conf/nips/XieS0L23}, and improving generalization~\cite{DBLP:conf/cvpr/JainSK0PM23}. Compared to input saliency and data attribution, model saliency is less explored. Weight sparsity~\cite{DBLP:conf/iclr/FrankleC19}, used in weight pruning, can be viewed as a weight saliency map that preserves model abilities. In NLP, model editing research~\cite{DBLP:conf/acl/DaiDHSCW22,DBLP:conf/nips/MengBAB22,DBLP:conf/iclr/MengSABB23} focuses on locating and modifying specific knowledge by targeting weights. This concept of an 'editable model region' aligns with weight saliency in NLP, where certain parameters are more influential and editable.

Compared with previous studies~\cite{DBLP:conf/nips/MengBAB22,DBLP:conf/iclr/MengSABB23,DBLP:conf/emnlp/YaoWT0LDC023} that are the most relevant to our work, a main difference should be highlighted. 
These approaches target at designing editing algorithms or evaluation paradigms to improve or assess the performance of model editing. However, previous work has not provided an in-depth analysis or an effective solution to preserve these abilities based on that analysis.
In contrast, our study seeks to analyze the factors for the degradation of the general abilities of the model in sequential editing and introduces the EAC framework to preserve these general abilities.
\section{Preliminary}
Model editing involves modifying the knowledge stored within LMs without requiring retraining, to better meet specific tasks or requirements. This process aims to refine various complex learned beliefs, including logical, spatial, and numerical knowledge.
In this paper, we study editing factual knowledge in the form of $(x_e, y_e)$\footnotemark{}.
\footnotetext{Can be also represented as knowledge triple \( t = (subject, relation, object) \).}
The language model $f_\theta \in \mathcal{F}$ can be defined as a function $f_\theta : \mathcal{X} \rightarrow \mathcal{Y}$, mapping input $x \in \mathcal{X}$ to its prediction $y \in \mathcal{Y}$. For an editing factual knowledge $(x_e, y_e)$, where $f_\theta(x_e) \neq y_e$, the goal of the model editing is to edit the parameters $\theta \in \Theta$ of the model $f_\theta$ to obtain an edited model $f_{\theta'}$, such that $f_{\theta'}(x_e) = y_e$.
In sequential editing, this process continues iteratively. Given a set of editing facts $\mathcal{E} = \left\{ (x_{ei}, y_{ei}) \mid i = 1, \ldots, n \right\}$ and an initial model $f_{\theta_{0}}$, each model editing step involves learning a function $K$ that produces an edited language model $f_{\theta_i}$ such that $K(f_{\theta_{i-1}}, (x_{ei}, y_{ei})) = f_{\theta_i}$.

The model editing process typically affects the predictions for a broad set of inputs closely related to the edited factual knowledge. This collection of inputs is referred to as the \textit{editing scope}. A successful edit should modify the model's behavior within the target scope while preserving performance on out-of-scope examples:
\[ 
f_{\theta_i}(x_{ei}) = 
\begin{cases} 
y_{ei} & \text{if} \,\, x_{ei} \in I(x_{ei}, y_{ei}), \\
f_{\theta_{i-1}}(x_{ei}) & \text{if} \,\, x_{ei} \in O(x_{ei}, y_{ei}). 
\end{cases} 
\]
The \textit{in-scope} \(I(x_{ei}, y_{ei})\) typically includes \(x_{ei}\) and its equivalence neighborhood \(N(x_{ei}, y_{ei})\), which encompasses related input/output pairs. In contrast, the \textit{out-of-scope} \(O(x_{ei}, y_{ei})\) comprises inputs unrelated to the edit example.
To evaluate the effectiveness of various model editing methods, previous works focus on evaluation along three dimensions: \emph{reliability}, \emph{generalization} and \emph{locality}~\cite{DBLP:conf/emnlp/CaoAT21,DBLP:conf/iclr/MitchellLBFM22,DBLP:conf/nips/MengBAB22,DBLP:conf/iclr/MengSABB23,DBLP:conf/emnlp/YaoWT0LDC023}. 

\section{Analysis of Ability Degradation}
\label{analysis}

ROME~\cite{DBLP:conf/nips/MengBAB22} and MEMIT~\cite{DBLP:conf/iclr/MengSABB23} are 
 currently popular model editing methods. Given that MEMIT builds upon the foundations of ROME by implementing residual distribution across multiple layers,  our analysis in the main text focuses primarily on ROME. 
This section presents a detailed analysis of how the model is affected during sequential editing using ROME. 
Statistical and visual analyses reveal that the degradation of general abilities is related to the unintentional introduction of the non-trivial noise that can make the parameter matrix after editing deviate from its original semantics space.

\subsection{Comparison with Fine-tuning Approach}
\begin{figure}[t]
  \subfigure[Editing-based methods]{
  \includegraphics[width=0.22\textwidth]{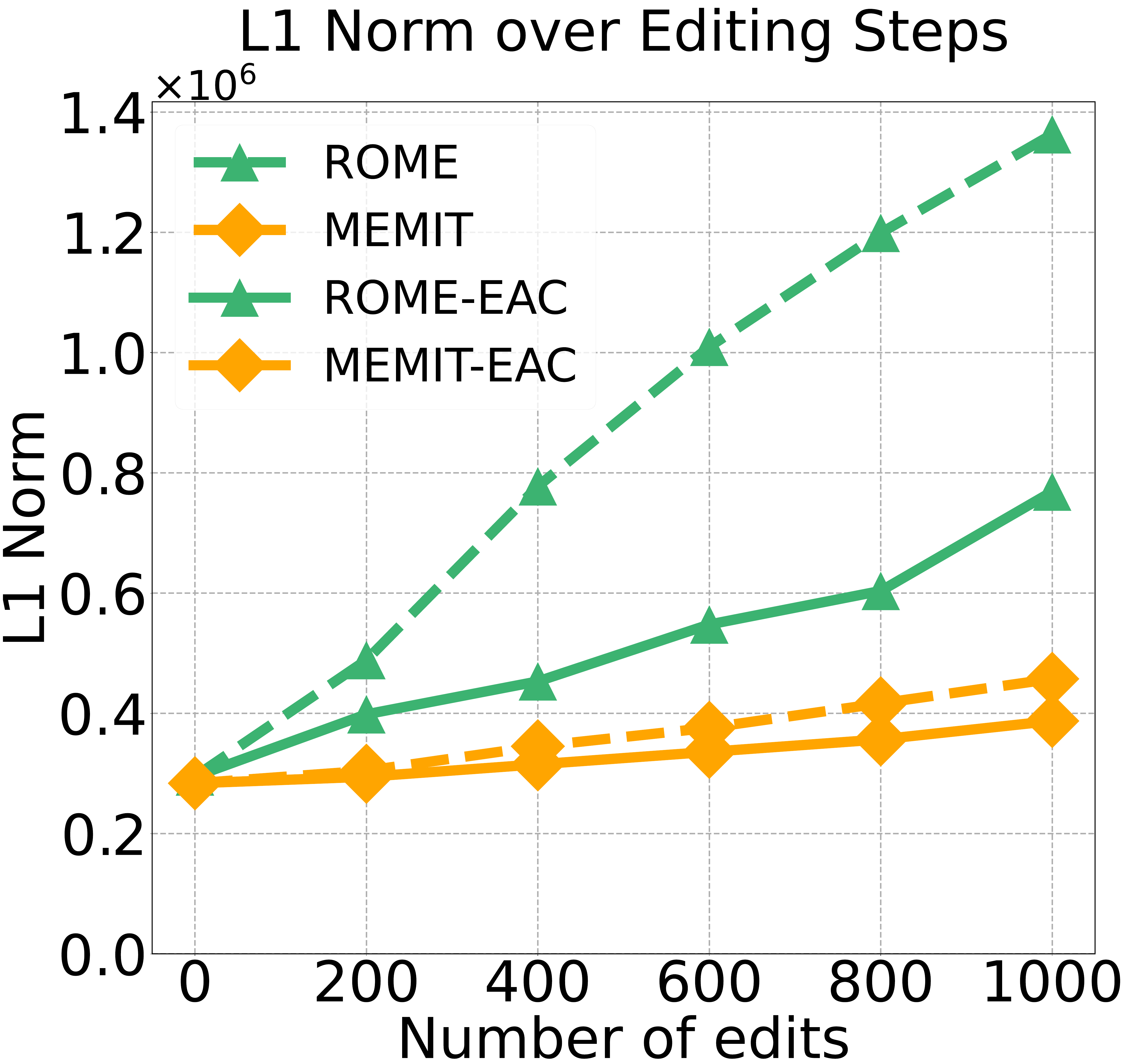}
  \label{fig-edit}}
  \subfigure[Fine-tuning approach]{
  \includegraphics[width=0.22\textwidth]{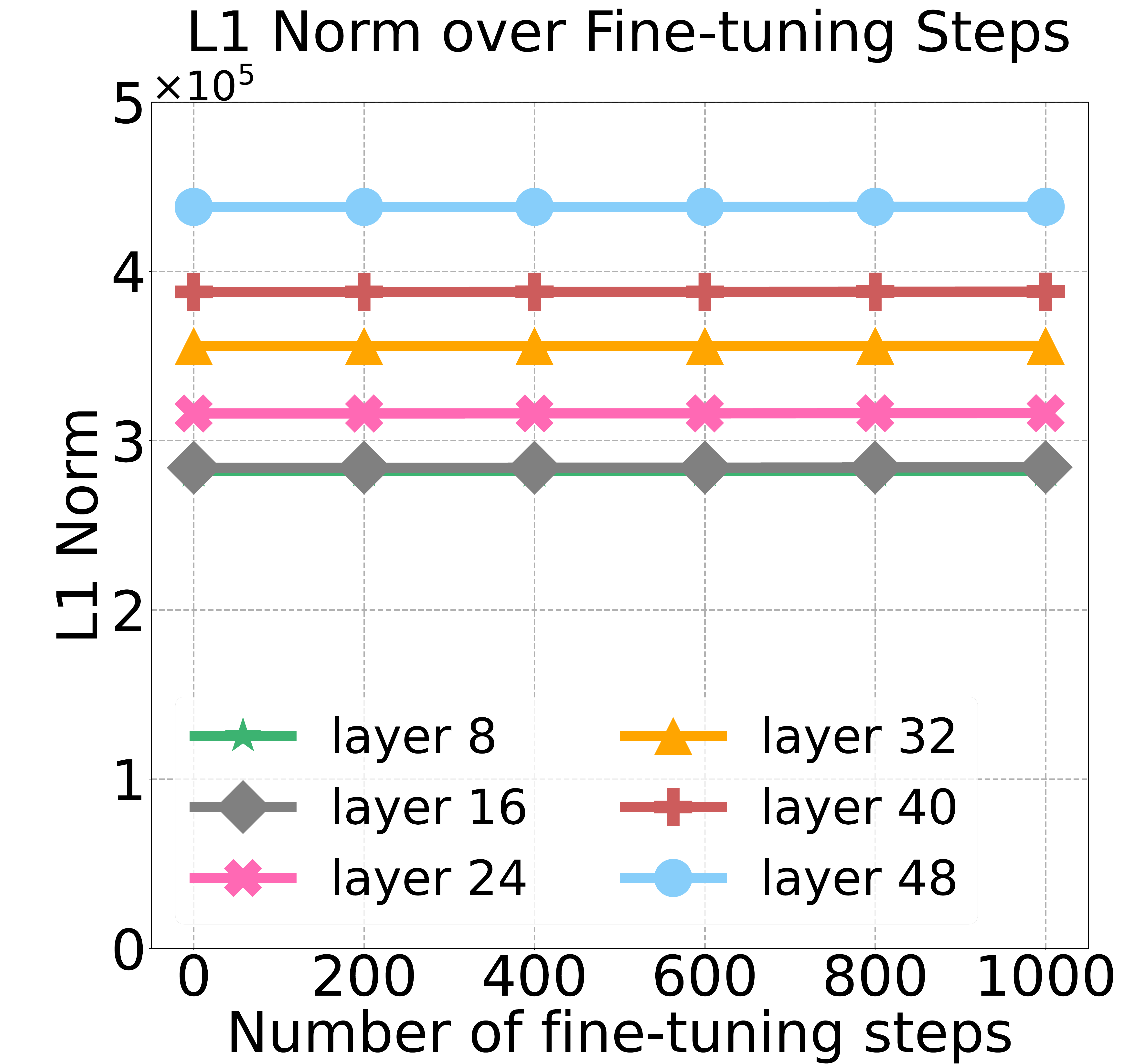}
  \label{fig-finetune}}
\vspace{-2mm}
\caption{Illustration of the change of L1 norm 
(a) in sequential editing at the edited layer using editing-based methods and 
(b) in fine-tuning different batch steps for selected layers.
Here we uniformly selected the layers of GPT2-XL for clarity when fine-tuning.} 
\vspace{-2mm}
\end{figure}

First, a statistical analysis is conducted by editing GPT2-XL~\cite{radford2019language} using the ZsRE~\cite{DBLP:conf/conll/LevySCZ17} dataset. 
Considering the L1 norm effectively quantifies the absolute changes in parameter values pre- and post-editing, while providing insights into feature weight distributions within the matrix, it is used to represent the degree of change in the parameter matrix.
As illustrated in Figure~\ref{fig-edit}, when using editing-based methods such as ROME and MEMIT, the L1 norm of the matrix at the edited layer increases significantly with the number of edits.
It can be seen that the norm increases by 317\% (ROME) and 61\% (MEMIT), respectively by the end of sequential editing.
This result highlights a significant deviation from the unedited model, emphasizing the impact of sequential edits on stability.

A gradient-based fine-tuning approach can markedly enhance the performance of the model on specific tasks while preserving its general abilities across other downstream tasks~\cite{DBLP:journals/corr/abs-2312-12740, DBLP:journals/corr/abs-2310-10047}. 
As depicted in Figure~\ref{fig-finetune}, there are no significant changes in the norm of the parameter matrix for the given layers, with a maximum change of only 0.27\%, even as the amount of fine-tuning knowledge increases. This stability in the parameter matrix norms suggests that the fine-tuning approach does not introduce significant non-trivial noise during the editing process. Thus, fine-tuning maintains the integrity and stability of the model's parameters, which is crucial for preserving its general abilities and preventing unintended non-trivial noise.

This comparison indicates that each editing method not only updates the intended fact as expected but also unintentionally introduces non-trivial noise into the model. This noise manifests as deviations in the parameter matrix during the sequential editing process. With each additional edit, the noise accumulates, progressively increasing the deviation in the parameter matrix. Consequently, as the number of edits grows, there is a significant deviation in the parameter matrix observed before and after the editing. This accumulated noise highlights the challenge of maintaining the stability and integrity of the parameter matrix through multiple edits, which can ultimately impact the general abilities of the model.

\begin{figure}[t]
  \subfigure{
  \includegraphics[width=0.45\textwidth]{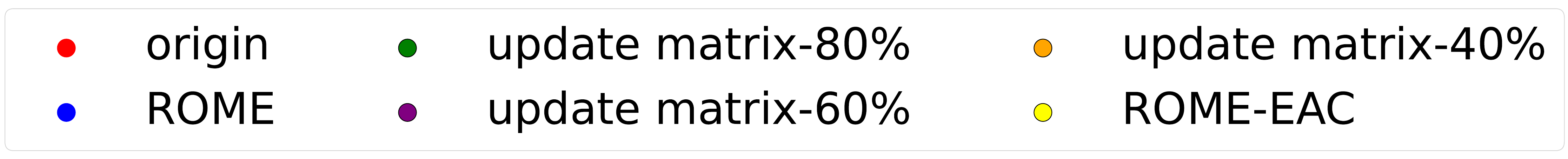}}
  \subfigure{
  \includegraphics[width=0.22\textwidth]{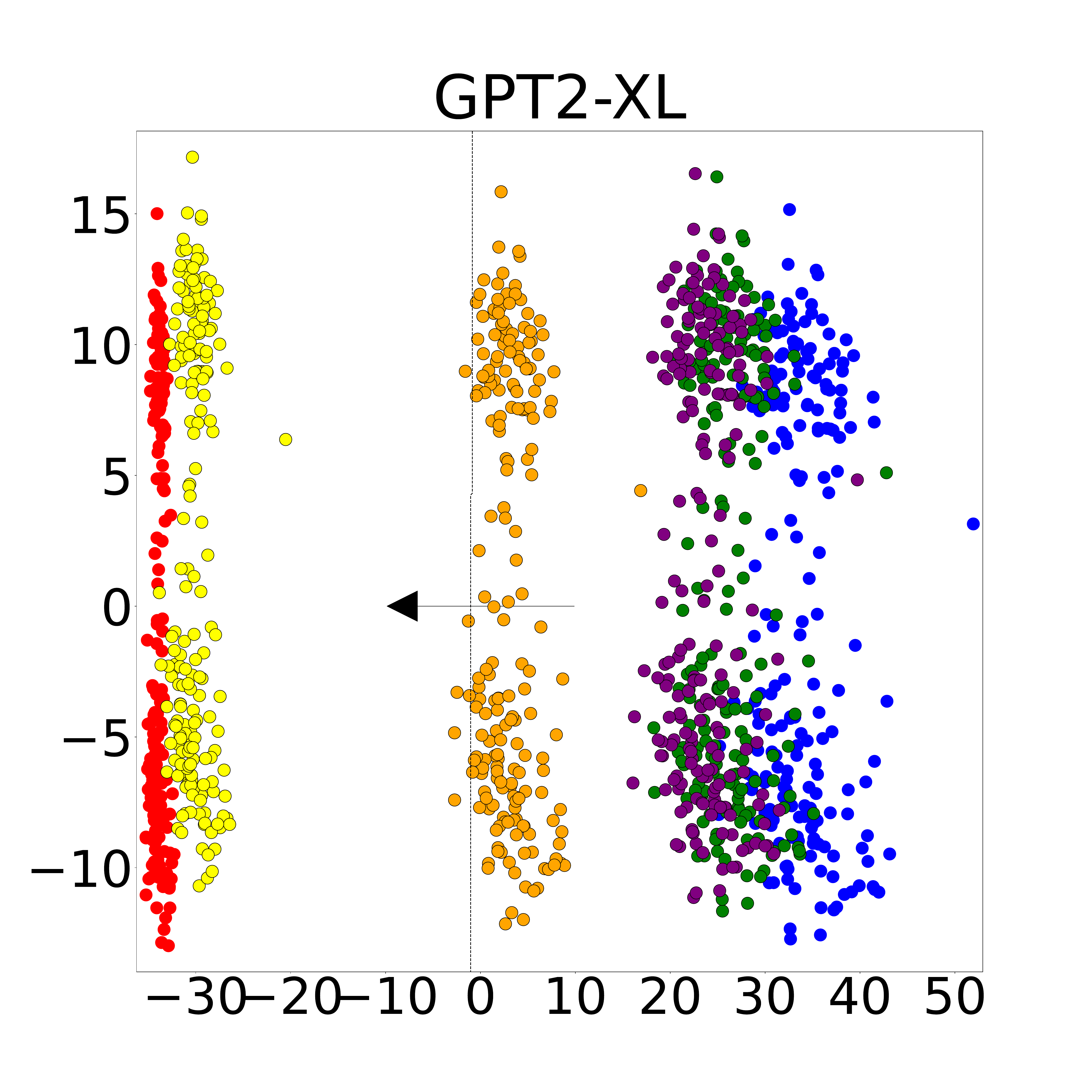}}
  \subfigure{
  \includegraphics[width=0.22\textwidth]{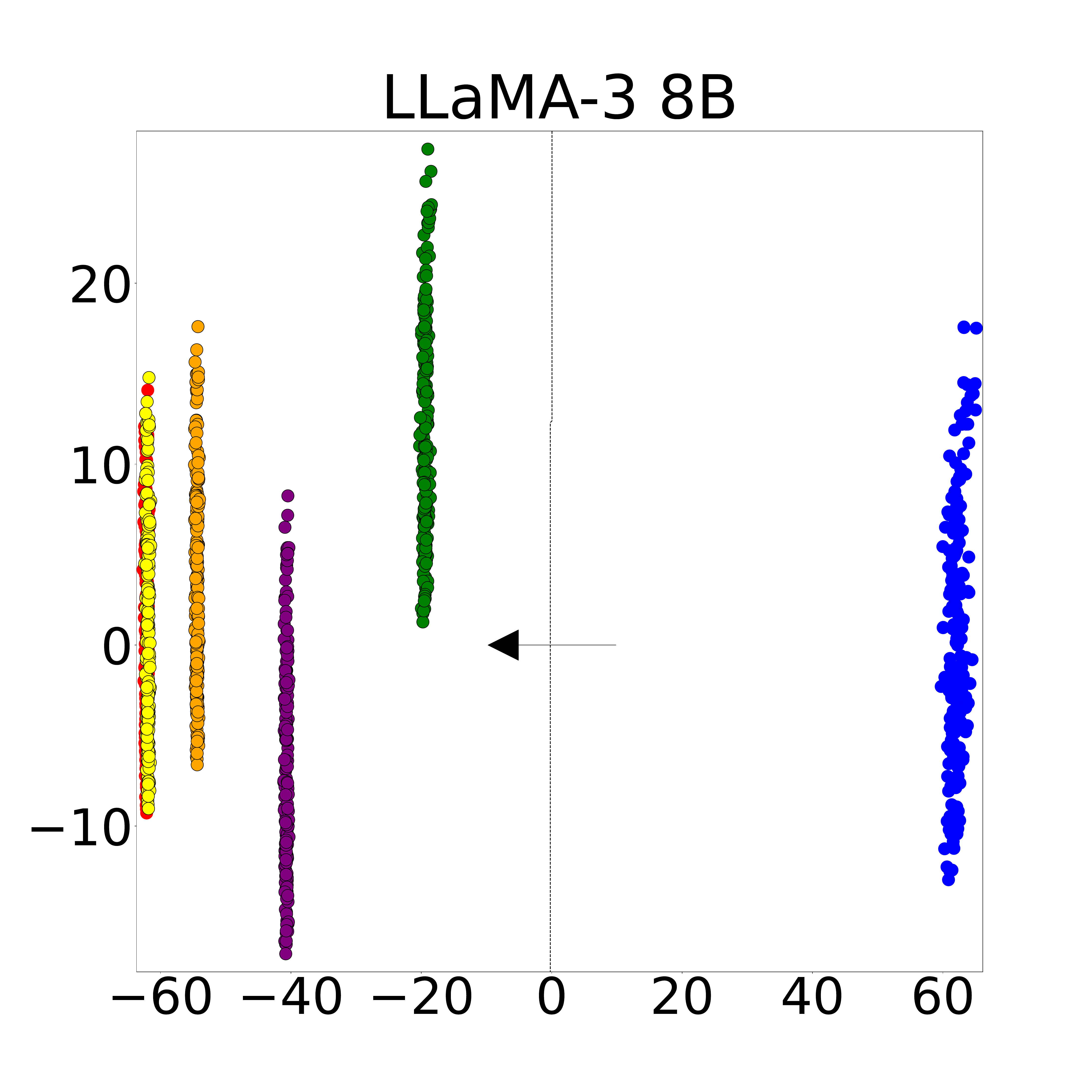}}
\vspace{-2mm}
\caption{Visialization of six sets of facts recalled by LLMs using 2-dimensional PCA. Note that this hidden state is also projected by a language modeling head (linear mapping) for next-token prediction, implying the linear structure in the corresponding representation space (the PCA assumption).} 
\vspace{-3mm}
\label{fig-pca}
\end{figure}

\subsection{Visualization Analysis} \label{sec_visual}
Following the ROME, the second layer of MLP \( W^{(l)}_{\text{proj}} \) is viewed as a linear associative memory~\cite{anderson1972simple, DBLP:journals/tc/Kohonen72}. This perspective observes that any linear operation \( W \) can operate as a key-value store for a set of vector keys \( K = [\mathbf{k}_1 | \mathbf{k}_2 | \dots] \) and corresponding vector values \( V = [\mathbf{v}_1 | \mathbf{v}_2 | \dots] \), by solving \( WK \approx V \)~\cite{DBLP:conf/nips/MengBAB22}.
The key-value pair \( (\mathbf{k}_i, \mathbf{v}_i) \) represents the representation of the input prompt, where $\mathbf{k}_i$ identifies patterns of the input and $\mathbf{v}_i$ is the fact recalled by the model, which is considered to gather all the information about how the model understands the prompt and how it will respond. By stacking $\mathbf{k}_i$ and $\mathbf{v}_i$ separately for each prompt, matrix \( K \) and \( V \) are obtained.
Based on this, 200 prompts of the same downstream task are collected to compute \( K \) and \( V \). 

On GPT2-XL and LLaMA-3 (8B), Principal Component Analysis is employed to visualize the hidden state of the facts of the downstream task recalled by the model. 
The first two principal components of six sets of facts, representing most features, are computed~\cite{zheng2024prompt}. Two of these are derived from recalling the model before editing and the model after editing without any constraint, respectively.
To explore the relationship between the deviation of the parameter matrix after editing and the resulting degradation of general abilities, four additional settings were tested by setting different percentages of the columns in the update matrix to zero, evenly distributed according to an arithmetic progression.

As illustrated in Figure~\ref{fig-pca}, the principal
components of facts recalled by the original model and the edited model without any constraint can be largely distinguished, whose boundaries (black dashed lines) can be easily fitted using logistic regression.
This indicates a significant semantic discrepancy between the facts recalled by the unconstrained edited model and the original model, explaining the decline in general abilities is related to matrix deviation.
Furthermore, when the deviation of the parameter matrix is constrained by reducing the norm of the update matrix, as shown by the black arrows, the principal components of the recalled facts by the edited model gradually align with those of the original model.
This shows that by reducing the norm of the update matrix, the deviation of the parameter matrix after editing can be constrained, making the semantic distribution of the model before and after editing similar, thereby preserving the general abilities of the edited model.
\section{EAC: \underline{E}diting \underline{A}nchor \underline{C}ompression}
\label{method}

\begin{figure}[t]
  \centering
  \includegraphics[width=0.48\textwidth]{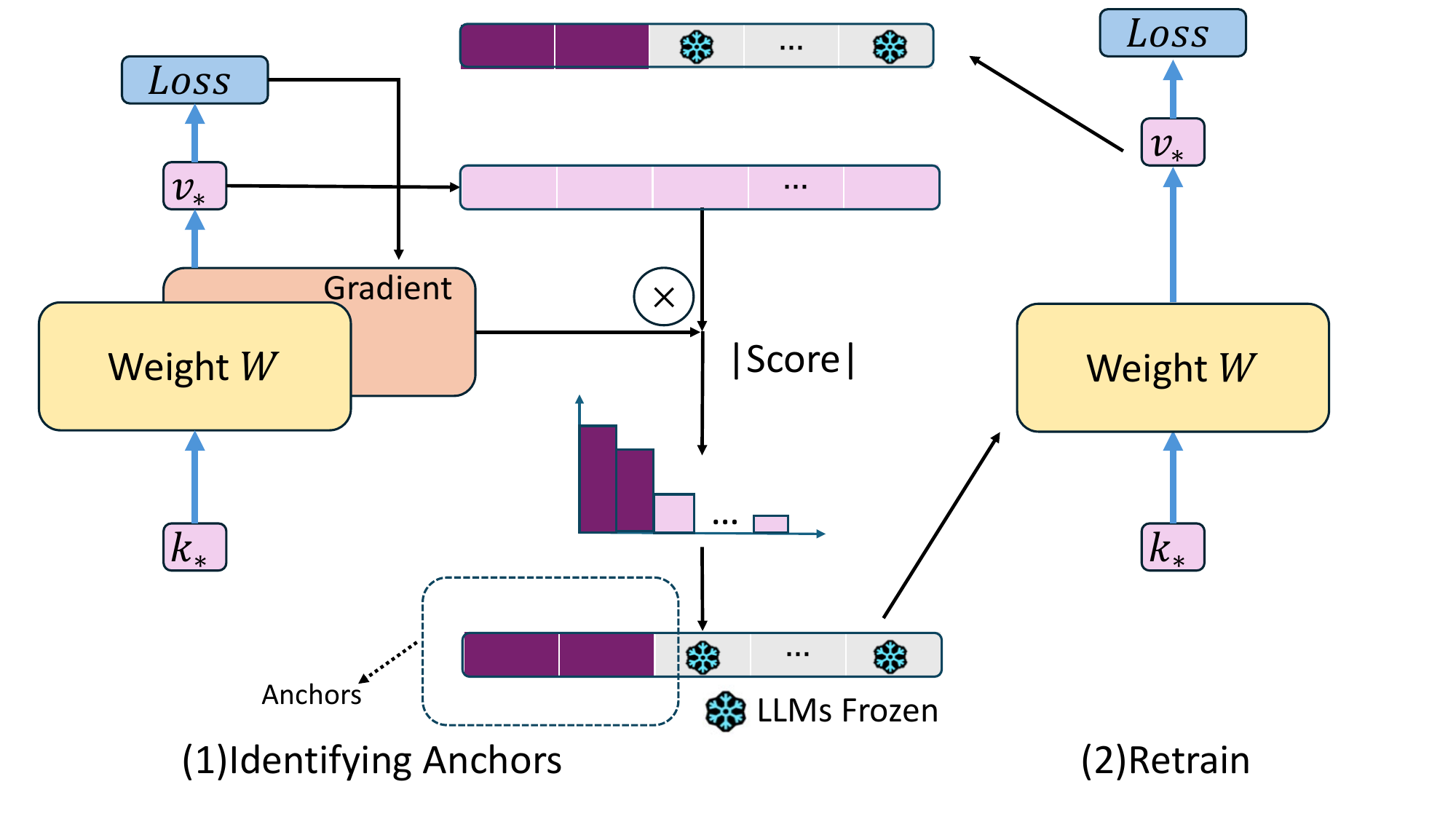}
  \vspace{-4mm}
  \caption{Proposed method: EAC. We first identify the key dimensions of the editing anchors using a weighted-gradient saliency map, followed by retraining on these dimensions to achieve the final optimization.}
  \vspace{-3mm}
  \label{EAC}
\end{figure}

In Section~\ref{analysis}, an in-depth analysis is provided on the factors that lead to the decrease in the general abilities of the model. 
Besides, it is found that the deviation of the edited parameter matrix could be constrained
by reducing the norm of the update matrix at each edit.
This helps maintain the semantic similarity of the facts recalled by the model before and after editing, ultimately preserving the general abilities of the edited model.
As depicted in Figure \ref{EAC}, a framework called EAC is proposed to compress information only in certain dimensions, thereby reducing the norm of the edited matrix and further constraining its deviation.

\subsection{Definition of Editing Anchors}
ROME uses an update matrix to insert a new knowledge triple \( t = (subject, relation, object) \). 
As mentioned in Section~\ref{sec_visual}, ROME calculates the update matrix by multiplying the pair \( (\mathbf{k}_*, \mathbf{v}_*) \), where \( \mathbf{k}_* \) identifies patterns of the input at the specified layer\footnotemark{} and \( \mathbf{v}_* \) is the fact recalled by the model.
Readers can refer to Appendix~\ref{b} for the details of ROME.
When injecting a new knowledge triple \( t = (subject, relation, object^*) \) to replace an old one \( t = (subject, relation, object) \), the specific part we aim to modify is the new relation \((relation, object^*) \), which is a property of the subject. It is believed that \( \mathbf{v}_* \) gathers all the information about how the model understands the subject and how it will respond thus we think that the new relation is primarily encoded in \( \mathbf{v}_* \).
Thus, EAC chooses to reduce the norm of \( \mathbf{v}_* \) for compression. Using a weighted gradient saliency map, EAC identifies high-scoring editing anchors crucial for encoding new relations. A scored elastic net is then applied to retrain and compress the editing information in key dimensions.
\footnotetext{Found by causal tracing methods~\cite{DBLP:conf/nips/MengBAB22}.}

\begin{figure*}[ht]
  \centering
  \includegraphics[width=0.5\textwidth]{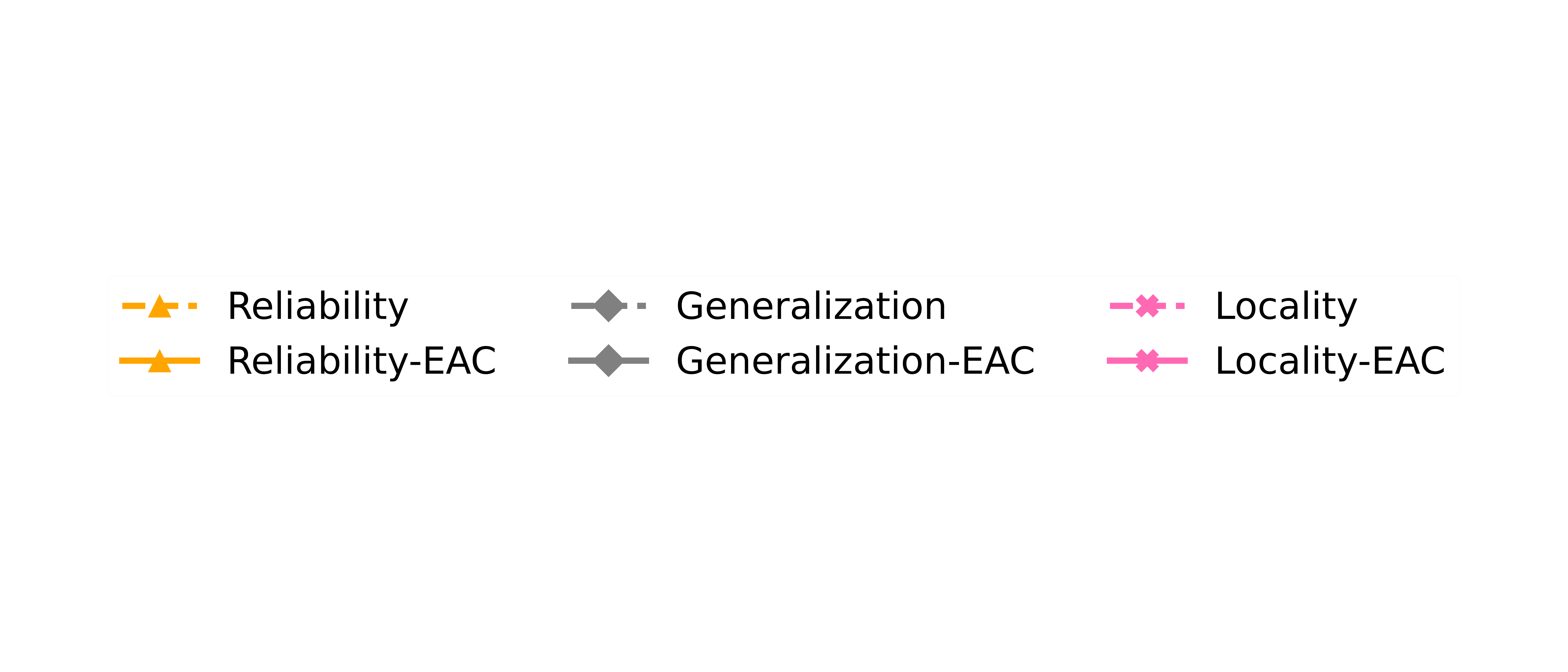}
  \vspace{-4mm}
\end{figure*}

\begin{figure*}[t]
  \centering
  \subfigure{
  \includegraphics[width=0.23\textwidth]{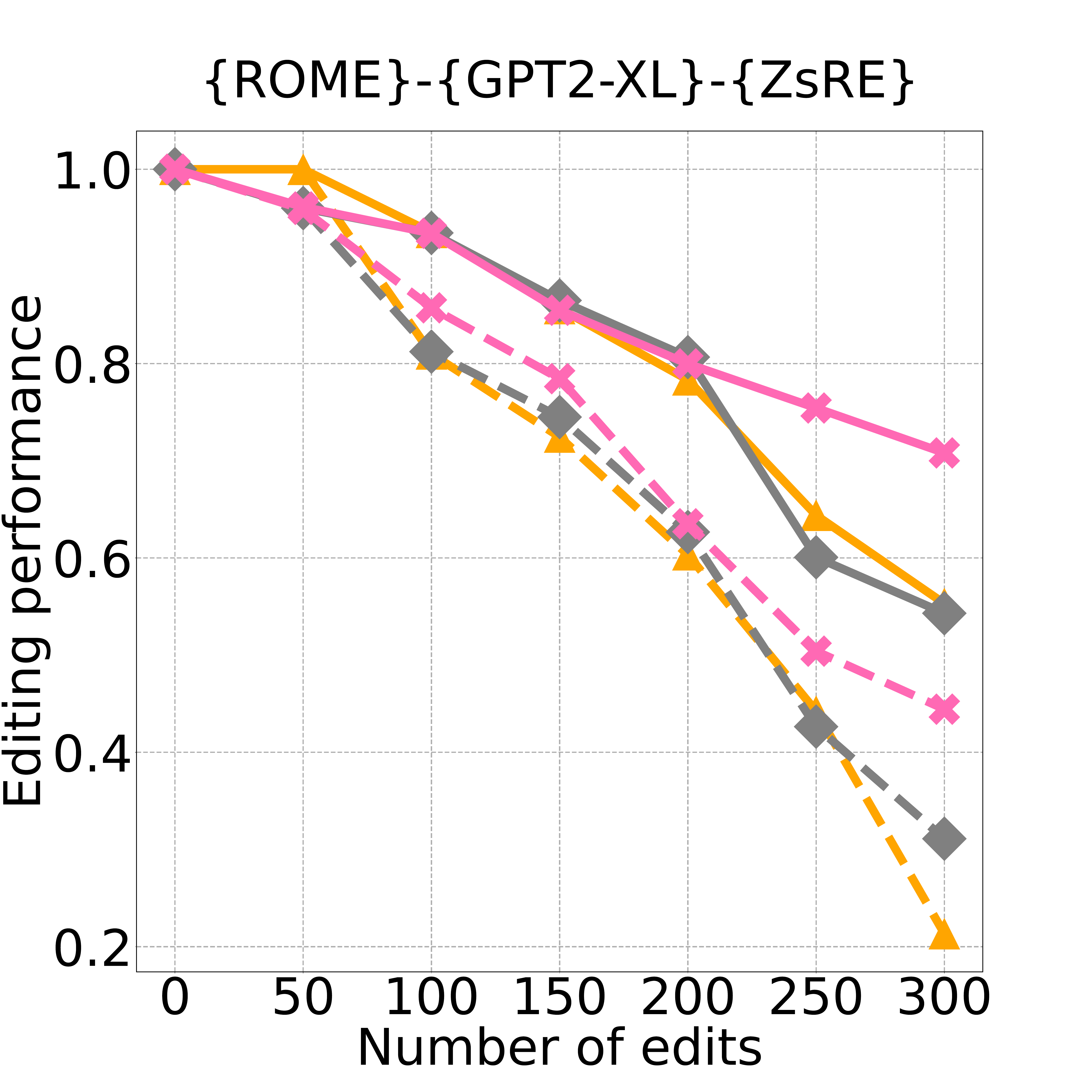}}
  \subfigure{
  \includegraphics[width=0.23\textwidth]{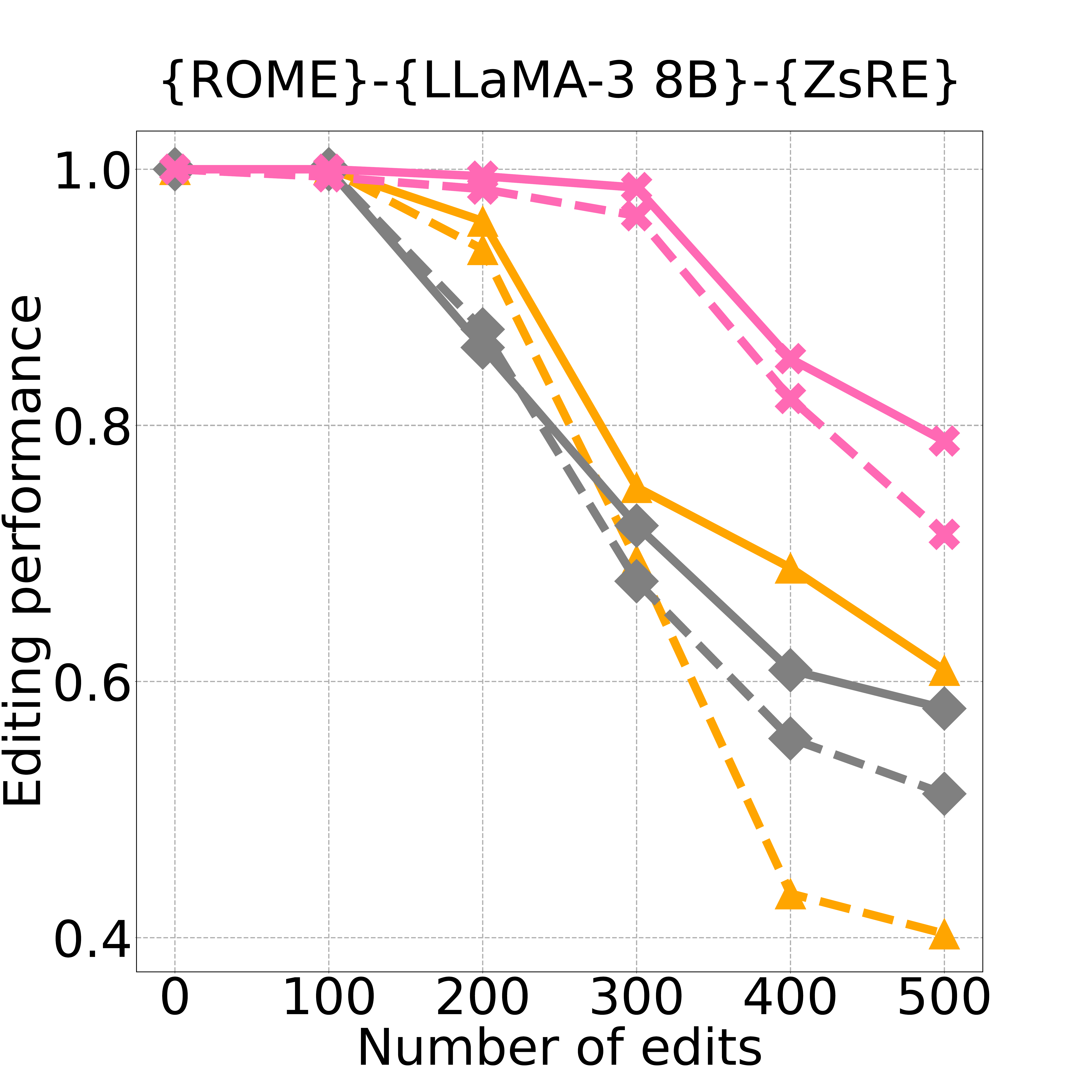}}
  \subfigure{
  \includegraphics[width=0.23\textwidth]{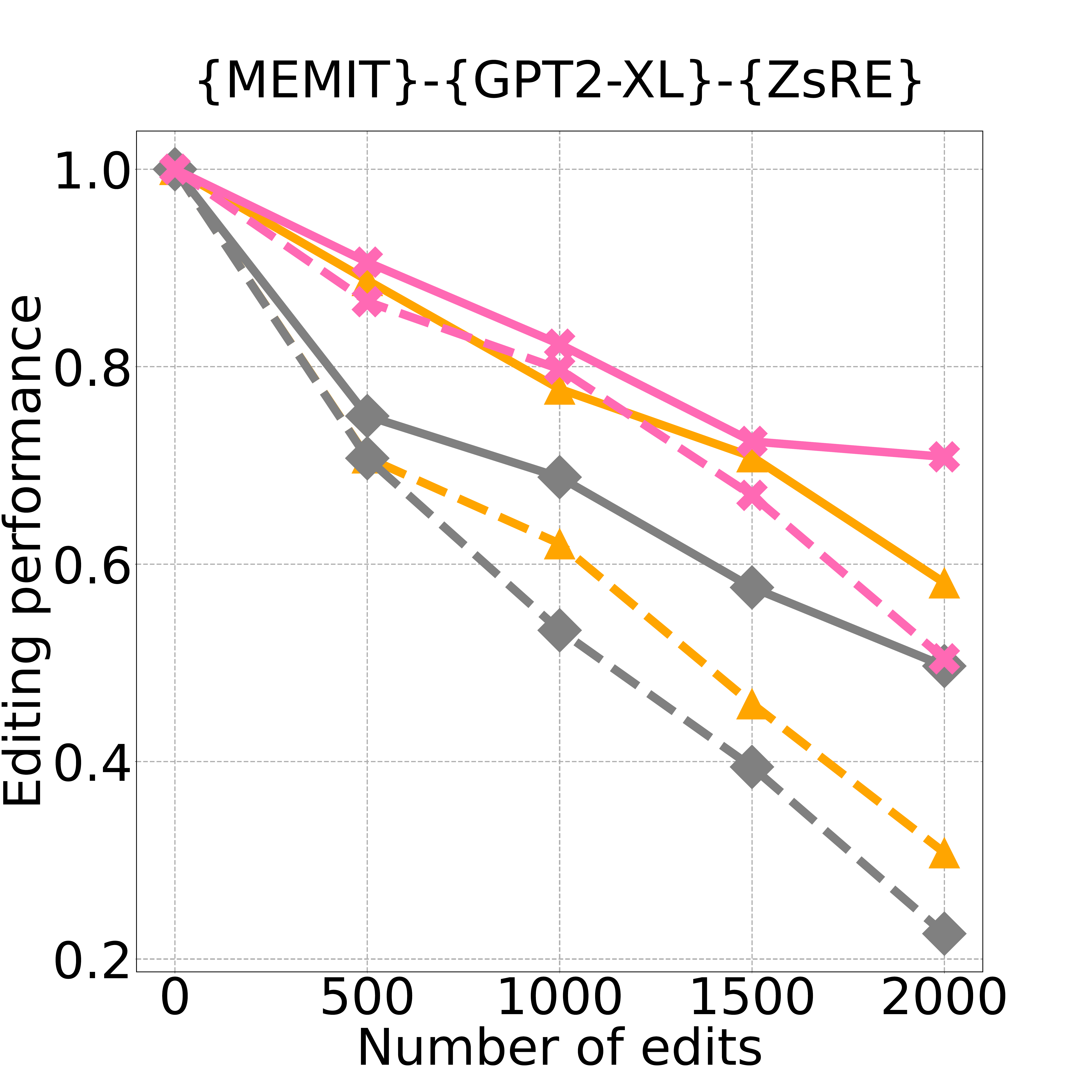}}
  \subfigure{
  \includegraphics[width=0.23\textwidth]{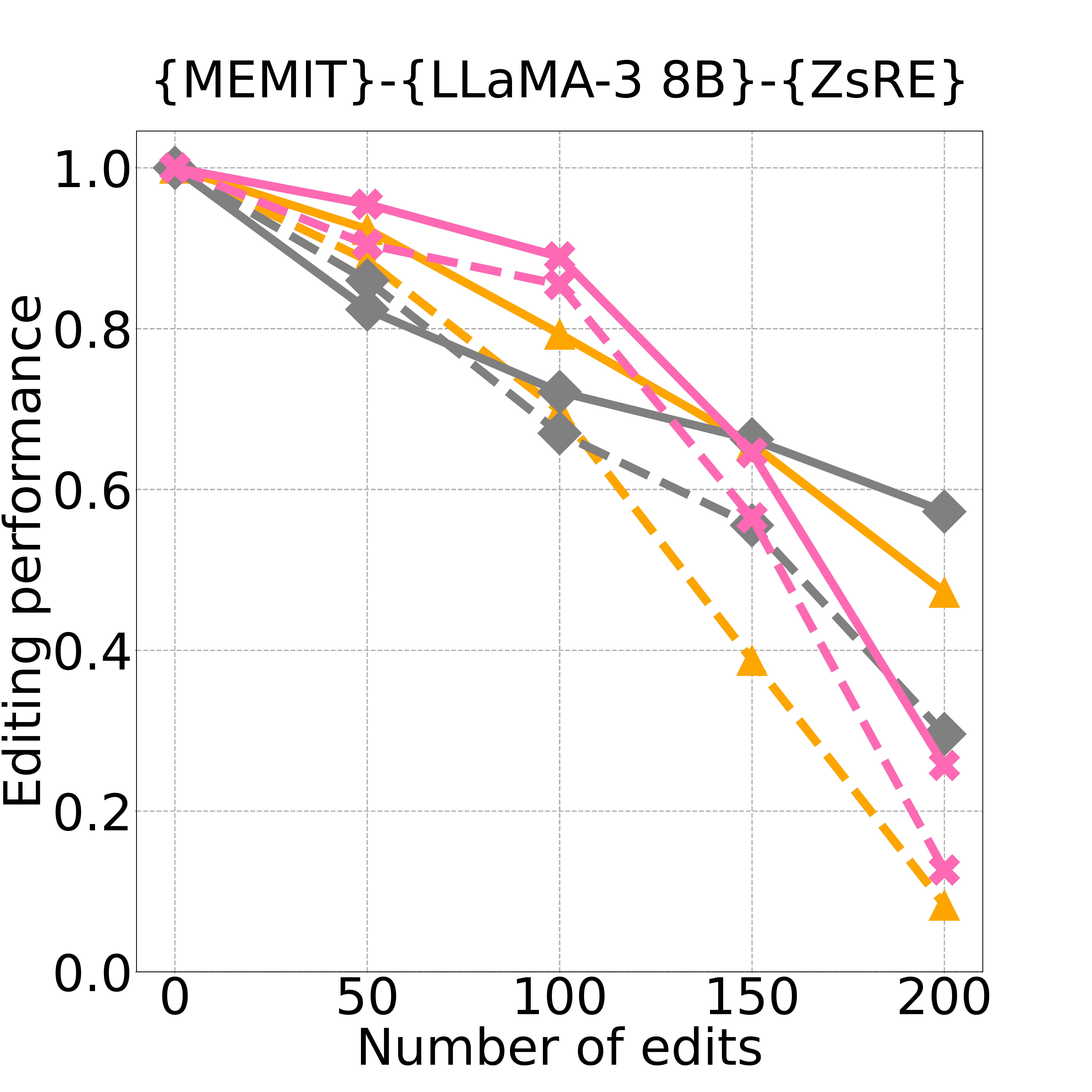}}
  \vspace{-2mm}
  \caption{Edited on the ZsRE dataset, the sequential editing performance of ROME and MEMIT with GPT2-XL and LLaMA-3 (8B) before and after the introduction of EAC, as the number of edits increases.}
  \vspace{-3mm}
  \label{edit-performance}
\end{figure*}

\subsection{Weighted-gradient Saliency Map}\label{a}
To reduce the norm of \( \mathbf{v}_* \) while preserving as much editing information as possible about the new relation, the goal is to compress this information over the smallest possible dimension. Drawing inspiration from gradient-based input saliency maps~\cite{DBLP:journals/corr/SmilkovTKVW17, DBLP:conf/nips/AdebayoGMGHK18}, a question is posed that whether a \textit{weight saliency map} can be constructed to aid compression. 
In previous work, ROME set \( \mathbf{v}_* = \arg\min_z \mathcal{L}(\mathbf{z}) \)~\cite{DBLP:conf/nips/MengBAB22}. Similar to the input saliency map, the gradient of this loss function with respect to each feature is utilized, and the magnitude of the values of \( \mathbf{v}_* \) is weighted accordingly to serve as the score for each feature: 
\begin{equation}
\text{score} = \mathbf{v}_* \odot \nabla \mathcal{L}(\mathbf{z}),
\label{score}
\end{equation}
where \(\odot\) is element-wise product. For GPT2-XL, the vectors $\mathbf{v}_*$ and $\mathbf{z}$ have dimensions of $1600 \times $1, whereas for LLaMA-3 (8B), the dimensions of these vectors are $4096 \times $1.
Based on Eq.~(\ref{score}), the desired weighted-gradient saliency map is obtained by applying a hard threshold:
\begin{equation}
\mathbf{m}_S = \mathbf{1} \left( \left| \text{score} \right| \geq \gamma \right),
\label{map}
\end{equation}
where \(\mathbf{1}(\mathbf{g} \geq \gamma)\) is an element-wise indicator function, which yields a value of 1 for the \(i\)-th element if \(g_i \geq \gamma\) and 0 otherwise.
\(| \cdot |\) is an element-wise absolute value operation, and \(\gamma > 0\) is a hard threshold. In practice, the value of \(\gamma\) is chosen according to different models and methods.
Specifically, the value of \(\gamma\) is chosen to retain the editing performance of the model in single editing. For more details refer to Appendix \ref{threshhold}.

With the introduction of the weighted gradient saliency map, the dimension of \( \mathbf{v}_* \) is split into two parts: one part represents the dimensions where the important editing anchors are located and it will be retrained to encode new relation, while the other part is set to 0, thereby reducing the norm of \( \mathbf{v}_* \). Based on Eq.~(\ref{map}), the \( \mathbf{v'}_* \), can be expressed as:
\begin{equation}
\mathbf{v'}_* \leftarrow \mathbf{m}_S \odot (\Delta \mathbf{v}_* + \mathbf{v}_*) + (\mathbf{1} - \mathbf{m}_S) \odot \mathbf{0},
\label{v}
\end{equation}
where \(\mathbf{1}\) denotes an all-one vector and \(\mathbf{0}\) denotes an all-zero vector. \(\Delta  \mathbf{v}_* \) is the part of \( \mathbf{v}_* \) that requires updating during retraining. Eq.~(\ref{v}) demonstrates that during retraining, only the dimensions where these important anchors are located need to be retrained to compress the editing information.

\subsection{Retraining Based on Scored Elastic Net}
After choosing important anchors, retraining for $\mathbf{v}_*$ is performed in this section.
To further compress the editing information, inspired by~\citet{zou2005regularization}, a score-based elastic net is also introduced during retraining:
\begin{equation}
\mathcal{L}_{0}(\mathbf{z}) = \lambda \|\mathbf{z}\|_{1, \alpha} + \mu \|\mathbf{z}\|_2^2,
\label{net}
\end{equation}
where \(\lambda \) and \(\mu \) are the hyper-parameters that control the strength of regularization and \( \mathbf{z} \) is the vector that causes the network to predict the target object in response to the factual prompt. Detailed hyper-parameters can be referred to in Appendix~\ref{hy}. Considering that the score computed in Eq.~(\ref{score}) represents the importance of the anchors for encoding the new relation, a weighted L1 norm is utilized when computing the L1 norm:
\begin{equation}
\|\mathbf{z}\|_{1, \alpha} = \sum_{i=1}^n \alpha_i |z_i|.
\end{equation}

In practice, we set \(\mathbf{\alpha} = \frac{1}{\text{score} + \epsilon}\), a small positive number \(\epsilon \) is introduced to prevent the score from being zero. 
Applying an elastic network, we ultimately derived the loss function during the retraining process to get the $\mathbf{v'}_*$ in Eq.~(\ref{v}):
\begin{equation}
\mathcal{L}_{r}(\mathbf{z}) =  \mathcal{L}(\mathbf{z}) + \mathcal{L}_{0}(\mathbf{z}).
\end{equation}

It is worth noting that when we make optimization here, only the dimensions where the editing anchors identified in section~\ref{a} are modified.
By introducing the elastic net, L1 regularization enables refining the selection of the editing anchors identified through the weighted-gradient saliency map during the retraining process. Meanwhile, L2 regularization effectively prevents model overfitting and improves the model's stability.
Finally, we complete the optimization. For specific optimization details, we recommend interested readers to refer to Appendix~\ref{b}.
Furthermore, the scored elastic net can also be applied to the FT. Readers can refer to Appendix \ref{FT} for more details.

\section{Experiments}

\begin{figure*}[ht]
  \centering
  \includegraphics[width=0.75\textwidth]{figures/legend.pdf}
  
  \subfigure{
  \includegraphics[width=0.23\textwidth]{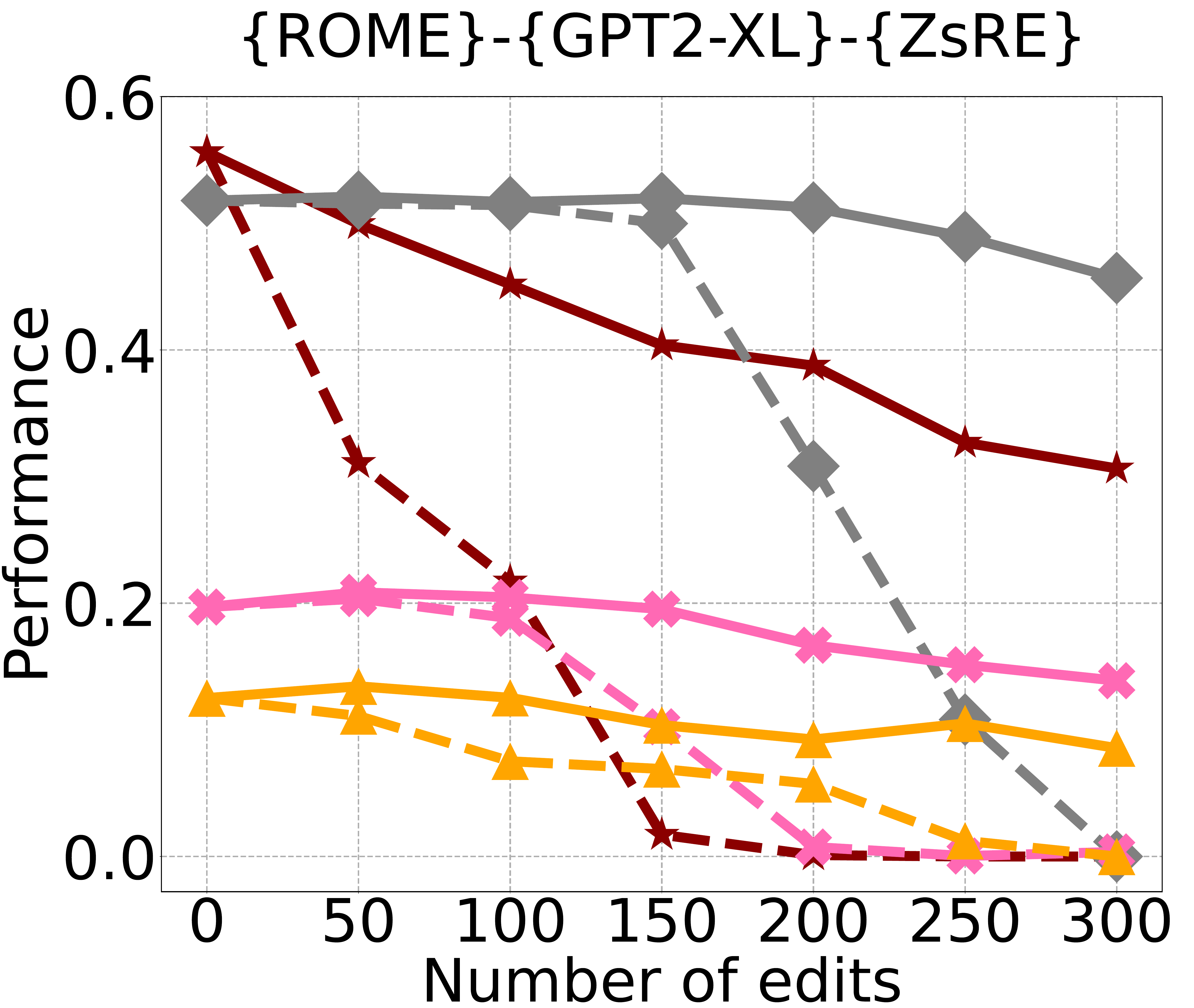}}
  \subfigure{
  \includegraphics[width=0.23\textwidth]{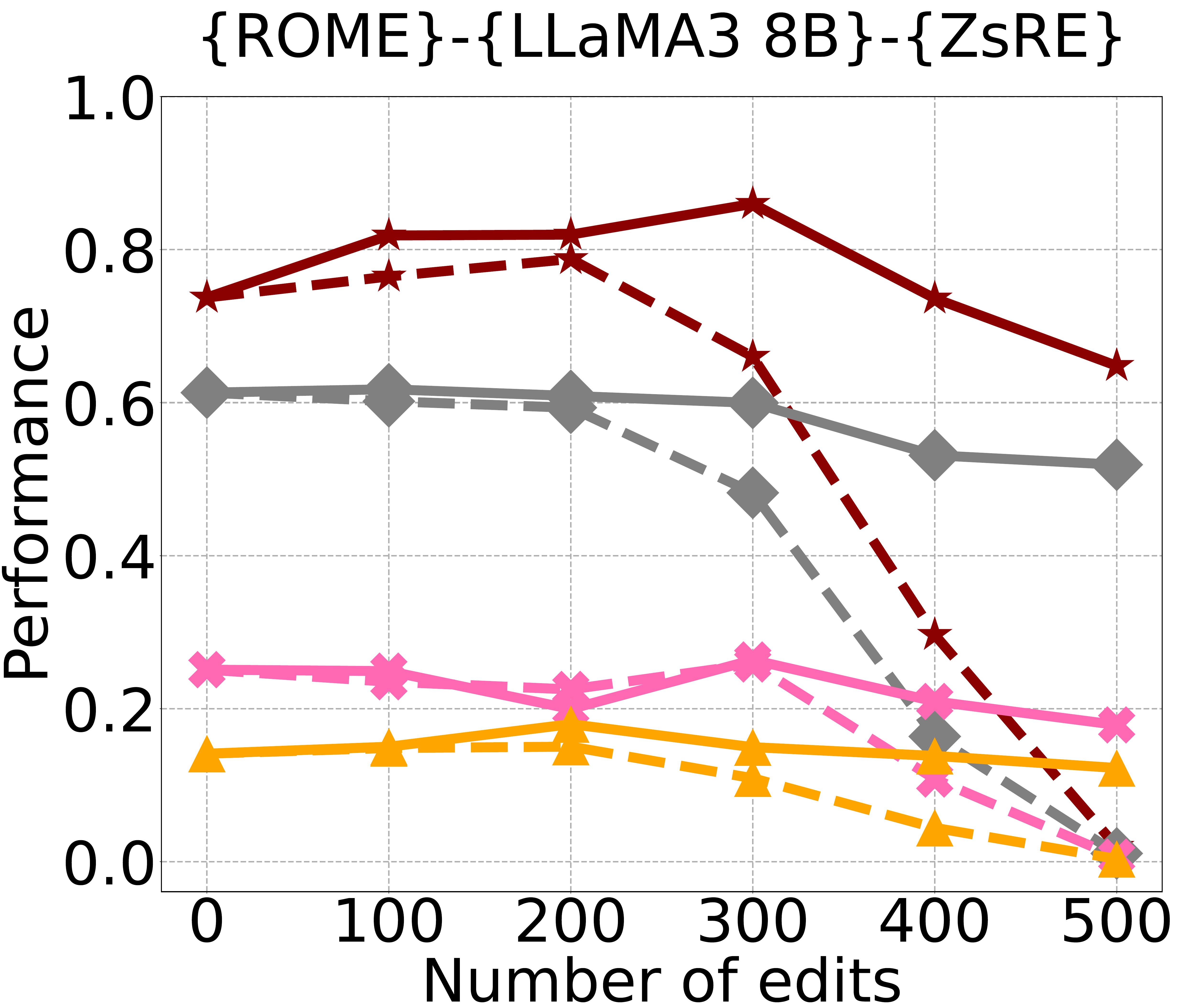}}
  \subfigure{
  \includegraphics[width=0.23\textwidth]{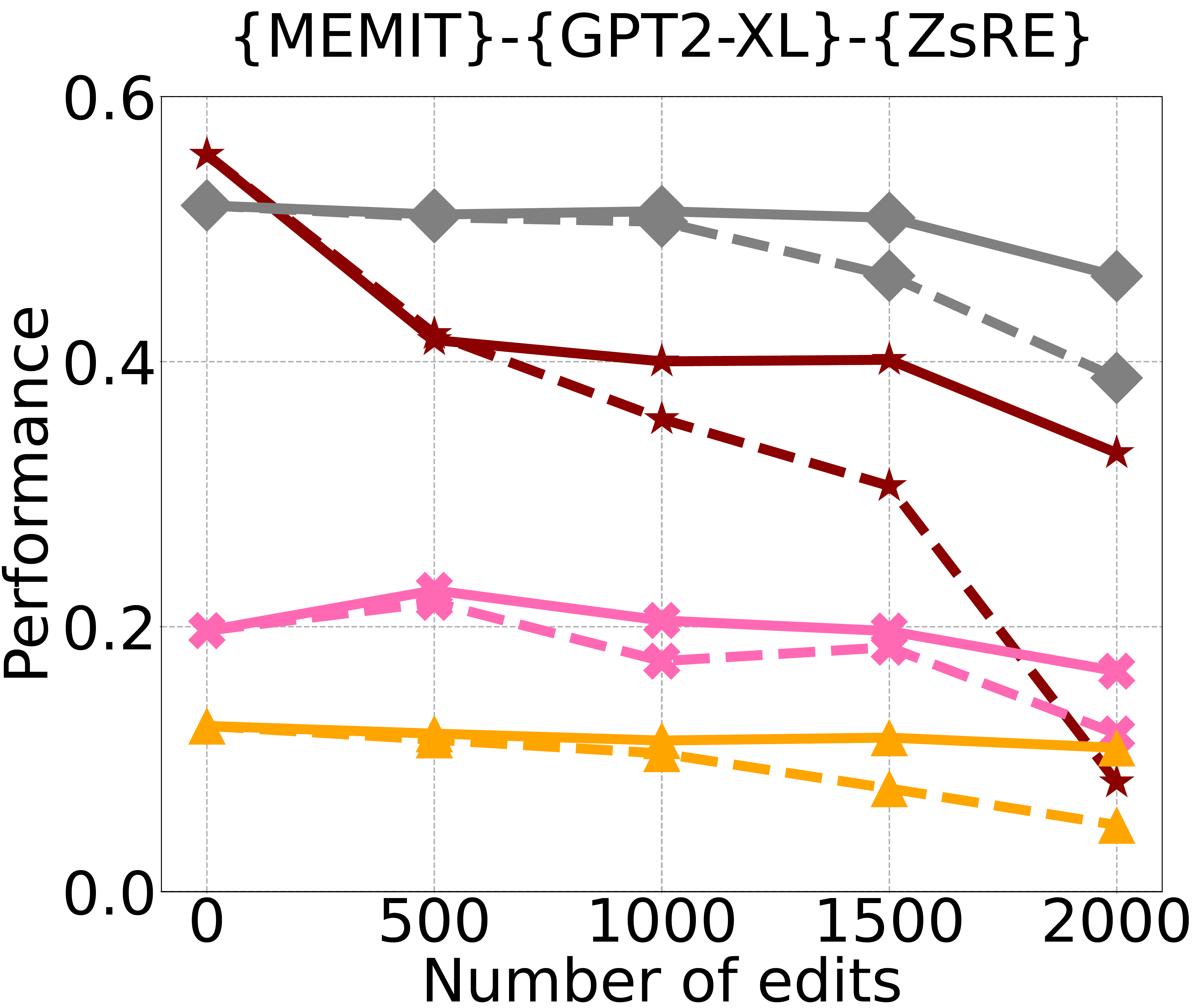}}
  \subfigure{
  \includegraphics[width=0.23\textwidth]{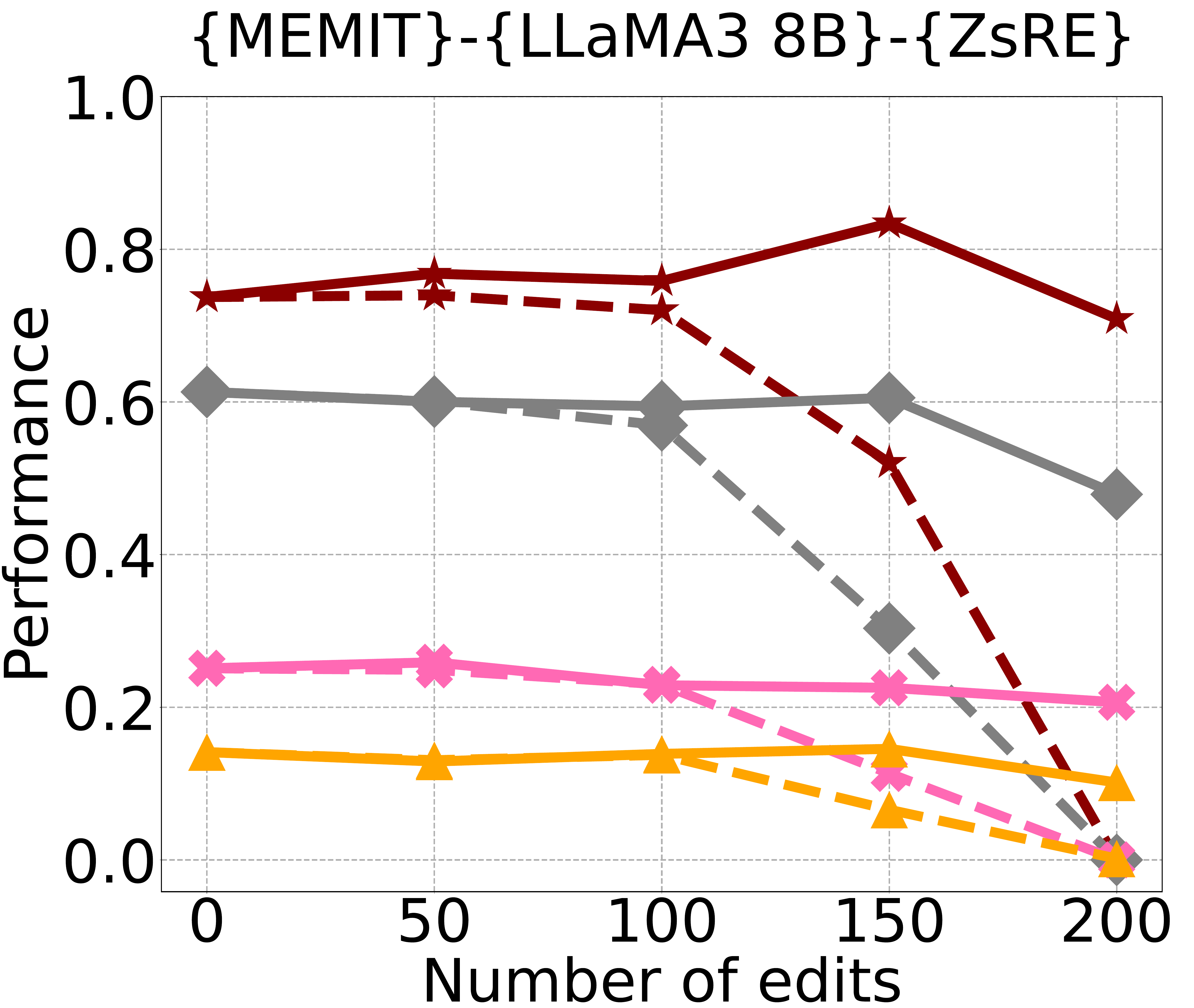}}
  \vspace{-2mm}
  \caption{Edited on the ZsRE dataset, the general task performance of ROME and MEMIT with GPT2-XL and LLaMA-3 (8B) before and after the introduction of EAC, as the number of edits increases.}
  \label{general ability}
  \vspace{-3mm}
\end{figure*}

\subsection{Experimental Setup}

Experiments were conducted on three LLMs, GPT2-XL~\cite{radford2019language}, LLaMA-3 (8B)~\cite{llama3} and LLaMA-2 (13B)~\cite{DBLP:journals/corr/abs-2307-09288},
using ROME~\cite{DBLP:conf/nips/MengBAB22} and MEMIT~\cite{DBLP:conf/iclr/MengSABB23} as baseline editing methods. 
The editing performance was evaluated on two datasets: ZsRE~\cite{DBLP:conf/conll/LevySCZ17} and CounterFact~\cite{DBLP:conf/nips/MengBAB22},
using reliability, generalization, and locality metrics~\cite{DBLP:conf/emnlp/CaoAT21,DBLP:conf/iclr/MitchellLBFM22,DBLP:conf/nips/MengBAB22,DBLP:conf/iclr/MengSABB23,DBLP:conf/emnlp/YaoWT0LDC023}.
Four downstream tasks were selected to measure the general abilities of models before and after editing:
\textbf{Natural language inference (NLI)}, 
\textbf{Open-domain QA}, \textbf{Summarization}
and \textbf{Sentiment analysis}.
Readers can refer to Appendix~\ref{detail} for more details.

\subsection{Main Results}

This section illustrates the editing performance and downstream task performance of edited models with GPT2-XL and LLaMA-3 (8B) on the ZsRE dataset. Due to page limitation, results of other LLMs and datasets were put in Appendix~\ref{app}

\paragraph{Editing Performance}
In previous work, the evaluation of editing performance has primarily focused on single editing scenarios, meaning that only the success of editing a single fact is assessed.
However, in sequential editing, the goal is for the model to retain all prior knowledge. To evaluate this, a set of sequential edits was applied, and the final model's reliability, generalization, and locality were assessed.
Applying ZsRE as the editing dataset, Figure~\ref{edit-performance} shows the sequential editing performance of ROME and MEMIT on GPT2-XL and LLaMA-3 (8B) before and after the introduction of EAC. 
The dashed line represents the ROME or MEMIT, while the solid line represents the ROME or MEMIT applying the EAC.
As sequential edits increase, models using ROME or MEMIT methods show a significant decline in reliability, generalization, and locality, retaining only partial knowledge from the latest edits while forgetting earlier information.
Besides, as shown in Figure~\ref{edit-performance}, the introduction of EAC brings improvements in the editing performance in sequential editing scenarios. By reducing the non-trivial noise introduced with each edit, EAC helps the model to retain the knowledge more effectively compared to previous methods.

\paragraph{General Abilities}
Applying ZsRE as the editing dataset, Figure~\ref{general ability} shows the performance on general tasks of ROME and MEMIT on GPT2-XL and LLaMA-3 (8B) before and after the introduction of EAC. 
The dashed line represents the ROME or MEMIT, while the solid line represents the ROME or MEMIT applying the EAC.
It can be seen that, when using the ROME or the MEMIT for sequential editing, the performance of the edited models on various tasks fluctuates significantly and shows a downward trend as the number of edits increases. 
After applying EAC, the general abilities of the model on downstream tasks are well preserved. However, the performance of the model on downstream tasks inevitably declined when the number of sequential edits was high. This indicates that some non-trivial noise is still introduced with each edit even when applying EAC. As these non-trivial noises accumulate, the general abilities of the model are compromised.

\begin{figure}[t]
  \centering
  \subfigure{
  \includegraphics[width=0.40\textwidth]{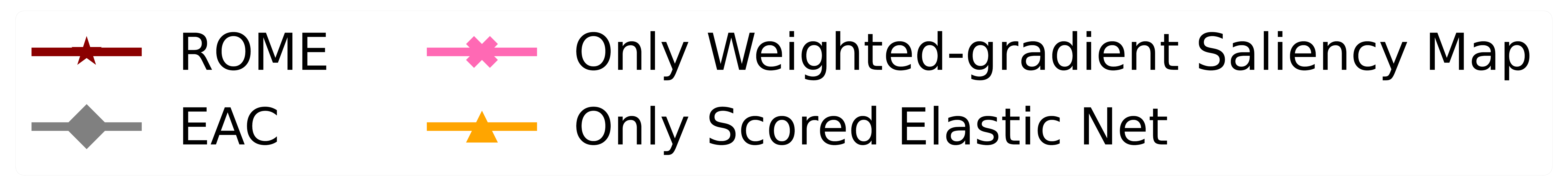}}
  \centering
  \subfigure{
  \includegraphics[width=0.22\textwidth]{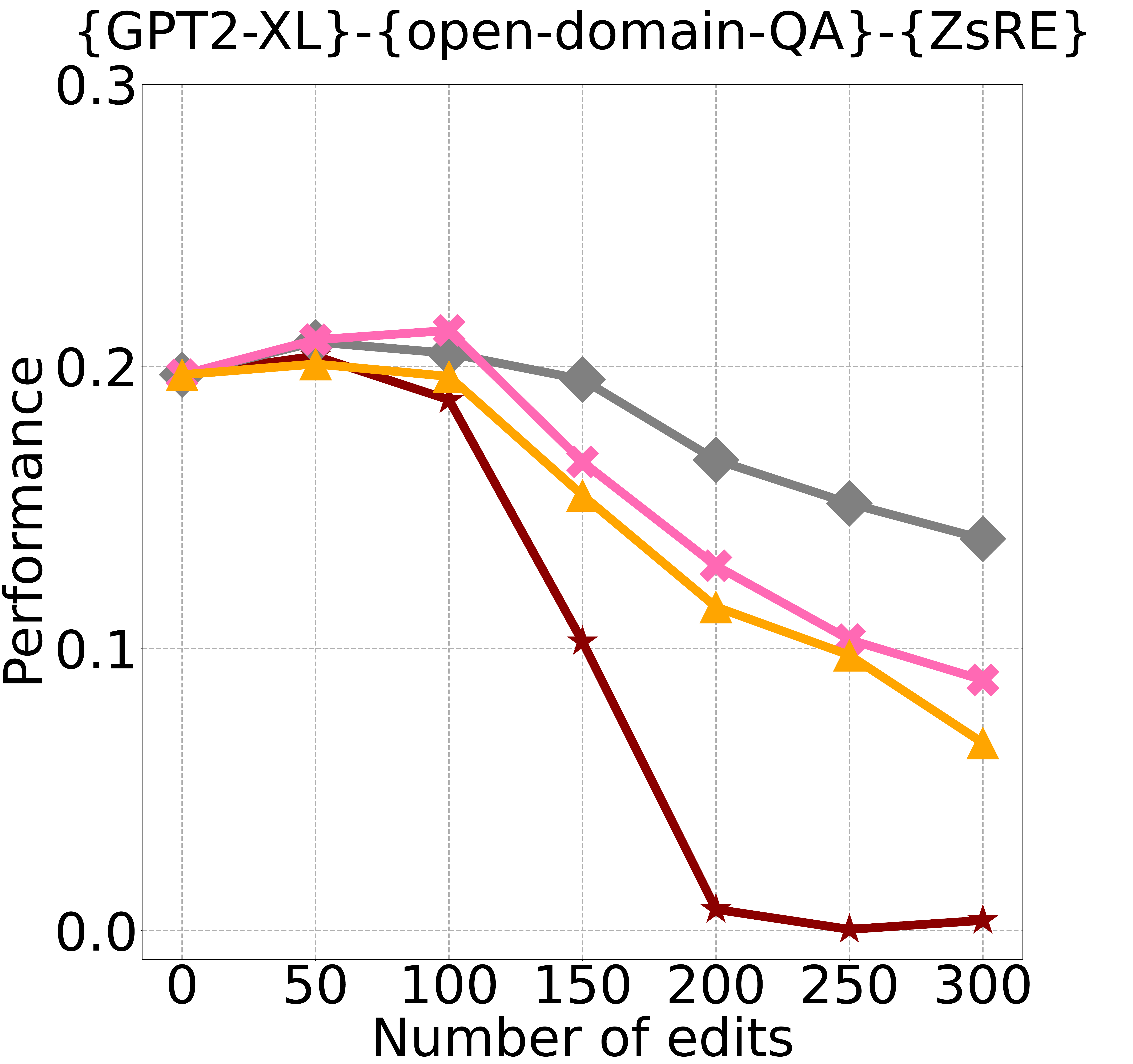}}
  \subfigure{
  \includegraphics[width=0.22\textwidth]{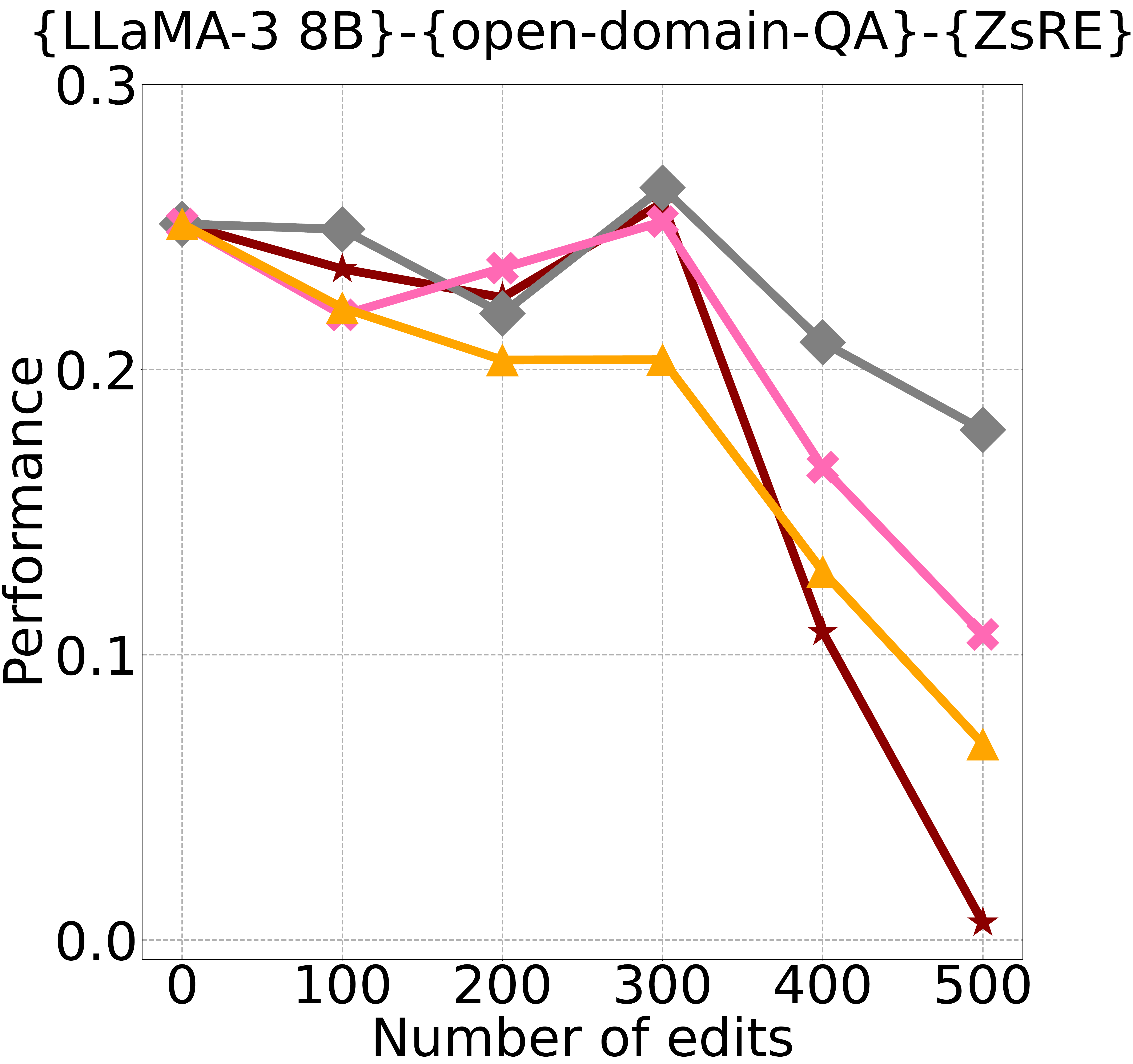}}
  \caption{Ablation analysis of performance on open-domain-QA for EAC. Results were conducted with GPT2-XL and LLaMA-3 (8B) on the ZsRE dataset.}
  \vspace{-1mm}
  \label{aba}
\end{figure}

\subsection{Ablation Study}
To validate the effectiveness of EAC, we performed ablation tests by removing either the weighted-gradient saliency map or the scored elastic net from EAC.
As depicted in Figure~\ref{aba}, we find that both the weighted-gradient saliency maps and the score-based elastic net retain a certain level of the general abilities of the model. It can also be seen that removing either the weighted-gradient saliency map or the scored elastic net results in a decreased ability to preserve the general abilities of the model compared to EAC, illustrating the effectiveness of EAC.
In addition, we observe a more significant decrease in the general abilities of the model when the weighted-gradient saliency map is removed. This suggests that the weighted-gradient saliency map plays a crucial role in EAC by effectively reducing the norm of the update matrix, which achieves this by setting some dimensions of \( \mathbf{v}_* \) to zero where the editing anchors are located, thereby constraining the deviation of the edited matrix and finally preserving the general abilities of the model.

\begin{table}[t]
    \centering
    \resizebox{0.85\linewidth}{!}{
    \begin{tabular}{lcc}
        \toprule
        & \textbf{GPT2-XL} & \textbf{LLaMA3-8B} \\
        \midrule
        ROME & 6.70s & 27.53s \\
        ROME-EAC & 6.21s & 23.74s \\
        \hline
        MEMIT & 5.23s & 11.73s \\
        MEMIT-EAC & 5.92s & 12.66s \\
        \bottomrule
    \end{tabular}
    }
    \caption{Comparison of editing time between the EAC framework and original methods}
    \vspace{-2mm}
    \label{time}
\end{table}

\subsection{Time Analysis}
In the EAC framework, the original training process for updating new knowledge is divided into two stages: first, selecting the important anchors, and second, retraining to complete the knowledge update.
Using GPT2-XL as an example, the original ROME method applies 20 optimization steps to update knowledge. In contrast, the EAC framework allocates 10 optimization steps to identify important editing anchors, followed by another 10 steps of retraining to finalize the knowledge update.
Additionally, since we reuse the gradient information generated during the selection of editing anchors, the EAC framework remains largely consistent with the previous method in terms of editing time and computational resource consumption, without incurring any extra computational overhead. This result is presented in Table~\ref{time}.

\begin{figure}[t]
  \centering
  \subfigure{
  \includegraphics[width=0.22\textwidth]{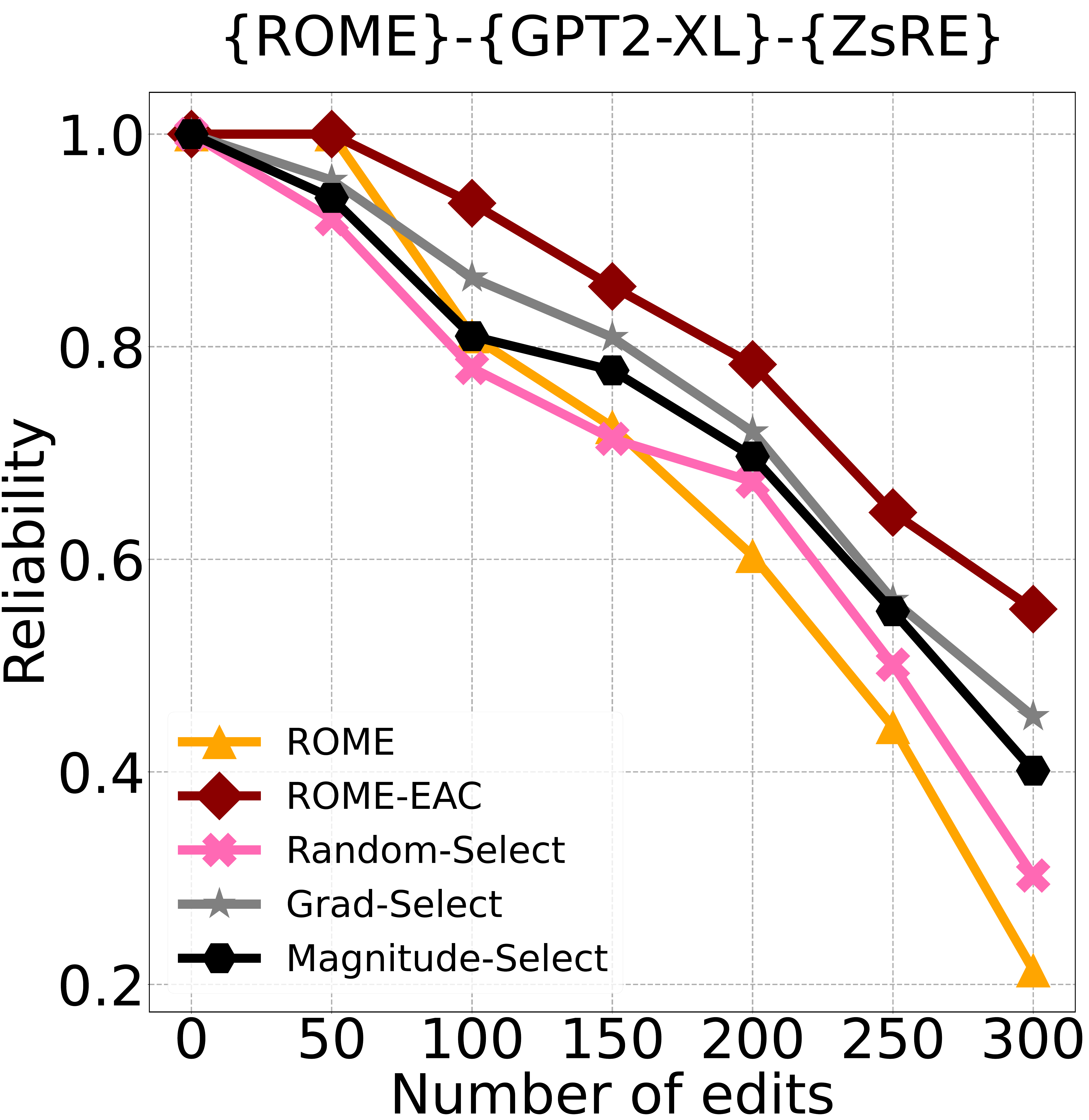}}
  \subfigure{
  \includegraphics[width=0.22\textwidth]{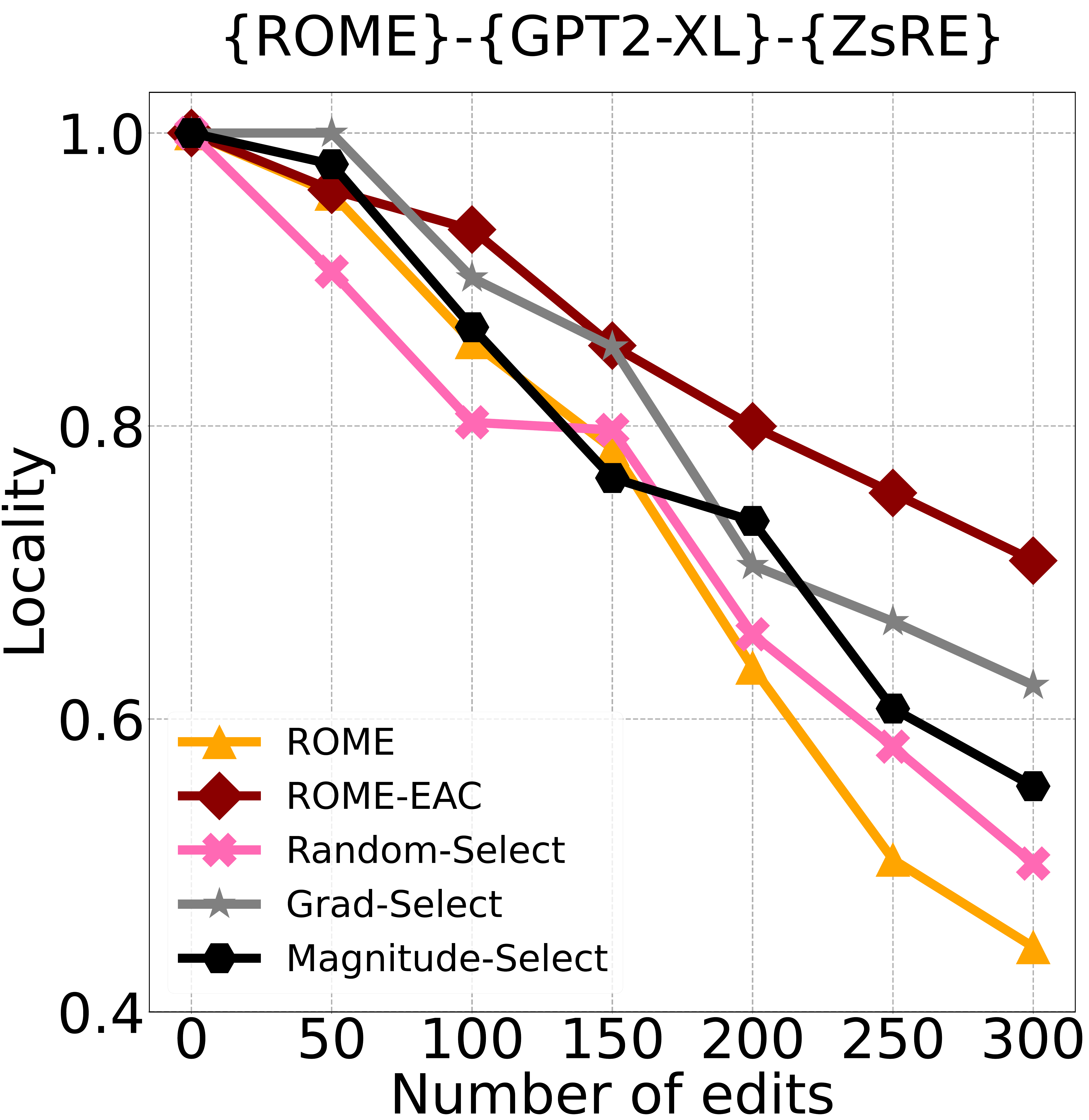}}
  \vspace{-2mm}
  \caption{Edited on the ZsRE dataset, the editing performance of different methods with GPT2-XL as the number of edits increases.}
  \label{case}
  \vspace{-3mm}
\end{figure}

\subsection{Editing Anchors Selecting Analysis}

In section \ref{method}, we employed a weighted-gradient saliency map to identify editing anchors, demonstrating the efficacy of this approach in subsequent experiments. Here, we delve deeper into the rationale behind using a weighted-gradient saliency map for determining these important anchors.
The weighted-gradient saliency map combines gradient sensitivity and vector magnitude to prioritize dimensions that significantly influence the knowledge vector, ensuring precise edits by focusing on areas both sensitive to changes and substantively important.
We validated this approach by comparing it to methods using random selection, gradients alone, or absolute values. Figure \ref{case} demonstrates that the weighted-gradient saliency map consistently outperforms other methods when applied to GPT2-XL on the ZsRE dataset. The superior performance of the weighted-gradient saliency map emphasizes its enhanced capability in precisely identifying key editing anchors within the model, ensuring greater accuracy and efficiency in the editing process.

\section{Conclusion}

This paper focuses on sequential model editing. Statistical and visual analyses reveal that each edit not only updates the desired fact but also introduces non-trivial noise, causing the model to deviate from its original semantic space. The accumulation of this noise negatively impacts the general abilities of LLMs.
To address this issue, a framework termed \textbf{E}diting \textbf{A}nchor \textbf{C}ompression (EAC) is proposed, which constrains edited matrix deviation by reducing the update matrix norm.
Experimental results show that EAC can effectively preserve the general abilities of edited models and the accuracy of editing facts in sequential editing.
For future work, we aim to investigate the impact of complex edits and integrate EAC with other editing methods.
 
\section*{Limitations}

Despite the effectiveness of EAC, our current studies still have limitations.
Firstly, similar to previous model editing research, we focus on factual knowledge assessment. However, it is also important to investigate whether editing other types of knowledge will affect general abilities and whether EAC is effective in this situation.
Secondly, our work focuses on sequential editing, where one fact is edited at a time. However, in real-world applications, we may need to change multiple facts in a single edit. Therefore, to enhance the scalability of model editing, batch-sequential settings should be emphasized in future studies.
Finally, to more fully validate the effectiveness of EAC, experiments should be conducted on larger-size models and across more downstream tasks.

\section*{Acknowledgements}

This work is funded by the National Science and Technology Major Project (No.2023ZD0121103).
We would like to express gratitude to the anonymous reviewers for their kind comments. 

\bibliography{custom}

\clearpage
\newpage
\appendix
\onecolumn
\section{Hard Threshold of EAC}\label{threshhold}
In constructing a weighted-gradient saliency map, the value of \(\gamma\) determines the number of the dimensions we select where important feature anchors are located. As the value of \(\gamma\) increases, the number of selected dimensions decreases, requiring the editing information to be compressed into a smaller space during the compression process. 
During compression, it is desired for the compression space to be as small as possible to preserve the general abilities of the model. However, reducing the compression space inevitably increases the loss of editing information, which reduces the editing performance of the model.
Therefore, to ensure editing performance in a single editing scenario, different values of \(\gamma\) are determined for various models, methods, and datasets. Fifty pieces of knowledge were randomly selected from the dataset, and reliability, generalization, and locality were measured after editing. The averages of these metrics were then taken as a measure of the editing performance of the model.
Table~\ref{value} presents the details of \(\gamma\), while Table~\ref{s} illustrates the corresponding editing performance before and after the introduction of EAC. $P_{x}$ denotes the value below which x\% of the values in the dataset.

\begin{table}[!htb]
\caption{The value of $\gamma$.}
\centering
\resizebox{0.45\textwidth}{!}{
\begin{tabular}{lcccc}
\toprule
\textbf{Datasets} & \textbf{Model} & \textbf{ROME} & \textbf{MEMIT} \\
\midrule
\multirow{2}{*}{\textbf{ZSRE}} & GPT-2 XL & $P_{80}$ & $P_{80}$ \\
 & LLaMA-3 (8B) & $P_{90}$ & $P_{95}$ \\
\midrule
\multirow{2}{*}{\textbf{COUNTERFACT}} & GPT-2 XL & $P_{85}$ & $P_{85}$ \\
 & LLaMA-3 (8B) & $P_{95}$ & $P_{95}$ \\
\bottomrule
\end{tabular}}
\label{value}
\end{table}

\begin{table}[!htb]
\caption{The value of $\gamma$.}
\centering
\resizebox{\textwidth}{!}{%
\begin{tabular}{lccccccccccccc}
\toprule
\multirow{1}{*}{Dataset} & \multirow{1}{*}{Method} & \multicolumn{3}{c}{\textbf{GPT-2 XL}} & \multicolumn{3}{c}{\textbf{LLaMA-3 (8B)}} \\
\cmidrule(lr){3-5} \cmidrule(lr){6-8}
& & \multicolumn{1}{c}{Reliability} & \multicolumn{1}{c}{Generalization} & \multicolumn{1}{c}{Locality} & \multicolumn{1}{c}{Reliability} & \multicolumn{1}{c}{Generalization} & \multicolumn{1}{c}{Locality} \\
\midrule
\multirow{1}{*}{ZsRE} & ROME & 1.0000 & 0.9112 & 0.9661 & 1.0000 & 0.9883 & 0.9600  \\
& ROME-EAC & 1.0000 & 0.8923 & 0.9560 & 0.9933 & 0.9733 & 0.9742  \\
\cmidrule(lr){2-8}
& MEMIT & 0.6928 & 0.5208 & 1.0000 & 0.9507 & 0.9333 & 0.9688  \\
& MEMIT-EAC & 0.6614 & 0.4968 & 0.9971 & 0.9503 & 0.9390 & 0.9767  \\
\midrule
\multirow{1}{*}{CounterFact} & ROME & 1.0000 & 0.4200 & 0.9600 & 1.0000 & 0.3600 & 0.7800  \\
& ROME-EAC & 0.9800 & 0.3800 & 0.9600 & 1.0000 & 0.3200 & 0.8800  \\
\cmidrule(lr){2-8}
& MEMIT & 0.9000 & 0.2200 & 1.0000 & 1.0000 & 0.3800 & 0.9500  \\
& MEMIT-EAC & 0.8000 & 0.1800 & 1.0000 & 1.0000 & 0.3200 & 0.9800  \\
\bottomrule
\end{tabular}%
}
\label{s}
\end{table}

\section{Optimization Details}\label{b}
ROME derives a closed-form solution to achieve the optimization:
\begin{equation}
\text{minimize} \ \| \widehat{W}K - V \| \ \text{such that} \ \widehat{W}\mathbf{k}_* = \mathbf{v}_* \ \text{by setting} \ \widehat{W} = W + \Lambda (C^{-1}\mathbf{k}_*)^T.
\end{equation}

Here \( W \) is the original matrix, \( C = KK^T \) is a constant that is pre-cached by estimating the uncentered covariance of \( \mathbf{k} \) from a sample of Wikipedia text, and \( \Lambda = (\mathbf{v}_* - W\mathbf{k}_*) / ( (C^{-1}\mathbf{k}_*)^T \mathbf{k}_*) \) is a vector proportional to the residual error of the new key-value pair on the original memory matrix.

In ROME, \(\mathbf{k}_*\) is derived from the following equation:
\begin{equation}
\mathbf{k}_* = \frac{1}{N} \sum_{j=1}^{N} \mathbf{k}(x_j + s), \quad \text{where} \quad \mathbf{k}(x) = \sigma \left( W_{f_c}^{(l^*)} \gamma \left( a_{[x],i}^{(l^*)} + h_{[x],i}^{(l^*-1)} \right) \right).
\end{equation}

ROME set $\mathbf{v}_* = \arg\min_z \mathcal{L}(z)$, where the objective $\mathcal{L}(z)$ is:
\begin{equation}
\frac{1}{N} \sum_{j=1}^{N} -\log \mathbb{P}_{G(m_{i}^{l^*}:=z))} \left[ o^* \mid x_j + p \right] + D_{KL} \left( \mathbb{P}_{G(m_{i}^{l^*}:=z)} \left[ x \mid p' \right] \parallel \mathbb{P}_{G} \left[ x \mid p' \right] \right).
\end{equation}

\section{Experimental Setup} \label{detail}

\subsection{Editing Methods}\label{EM}

In our experiments, Two popular editing methods including ROME and MEMIT were selected as baselines.

\textbf{ROME} \cite{DBLP:conf/nips/MengBAB22}: it first localized the factual knowledge at a specific layer in the transformer MLP modules, and then updated the knowledge by directly writing new key-value pairs in the MLP module.

\textbf{MEMIT} \cite{DBLP:conf/iclr/MengSABB23}: it extended ROME to edit a large set of facts and updated a set of MLP layers to update knowledge.

The ability of these methods was assessed based on EasyEdit~\cite{DBLP:journals/corr/abs-2308-07269}, an easy-to-use knowledge editing framework which integrates the released codes and hyperparameters from previous methods.

\subsection{Editing Datasets}\label{dat}
In our experiment, two popular model editing datasets \textsc{ZsRE}~\cite{DBLP:conf/conll/LevySCZ17} and \textsc{CounterFact}~\cite{DBLP:conf/nips/MengBAB22} were adopted.

\textbf{\textsc{ZsRE}} is a QA dataset using question rephrasings generated by back-translation as the equivalence neighborhood.
Each input is a question about an entity, and plausible alternative edit labels are sampled from the top-ranked predictions of a BART-base model trained on \textsc{ZsRE}.

\textbf{\textsc{CounterFact}} accounts for counterfacts that start with low scores in comparison to correct facts. It constructs out-of-scope data by substituting the subject entity for a proximate subject entity sharing a predicate. This alteration enables us to differentiate between superficial wording changes and more significant modifications that correspond to a meaningful shift in a fact. 

\subsection{Metrics for Evaluating Editing Performance}\label{Mediting performance}
\paragraph{Reliability} means that given an editing factual knowledge, the edited model should produce the expected predictions. The reliability is measured as the average accuracy on the edit case:
\begin{equation}
\mathbb{E}_{(x'_{ei}, y'_{ei}) \sim \{(x_{ei}, y_{ei})\}} \mathbf{1} \left\{ \arg\max_y f_{\theta_{i}} \left( y \mid x'_{ei} \right) = y'_{ei} \right\}.
\label{rel}
\end{equation}

\paragraph{Generalization} means that edited models should be able to recall the updated knowledge when prompted within the editing scope. The generalization is assessed by the average accuracy of the model on examples uniformly sampled from the equivalence neighborhood:
\begin{equation}
\mathbb{E}_{(x'_{ei}, y'_{ei}) \sim N(x_{ei}, y_{ei})} \mathbf{1} \left\{ \arg\max_y f_{\theta_{i}} \left( y \mid x'_{ei} \right) = y'_{ei} \right\}.
\label{gen}
\end{equation}

\paragraph{Locality} means that the edited model should remain unchanged in response to prompts that are irrelevant or the out-of-scope. The locality is evaluated by the rate at which the edited model's predictions remain unchanged compared to the pre-edit model.
\begin{equation}
\mathbb{E}_{(x'_{ei}, y'_{ei}) \sim O(x_{ei}, y_{ei})} \mathbf{1} \left\{ f_{\theta_{i}} \left( y \mid x'_{ei} \right) = f_{\theta_{i-1}} \left( y \mid x'_{ei} \right) \right\}.
\label{loc}
\end{equation}

\subsection{Downstream Tasks}\label{pro}

Four downstream tasks were selected to measure the general abilities of models before and after editing:
\textbf{Natural language inference (NLI)} on the RTE~\cite{DBLP:conf/mlcw/DaganGM05}, and the results were measured by accuracy of two-way classification.
\textbf{Open-domain QA} on the Natural Question~\cite{DBLP:journals/tacl/KwiatkowskiPRCP19}, and the results were measured by exact match (EM) with the reference answer after minor normalization as in \citet{DBLP:conf/acl/ChenFWB17} and \citet{DBLP:conf/acl/LeeCT19}.
\textbf{Summarization} on the SAMSum~\cite{gliwa-etal-2019-samsum}, and the results were measured by the average of ROUGE-1, ROUGE-2 and ROUGE-L as in \citet{lin-2004-rouge}.
\textbf{Sentiment analysis} on the SST2~\cite{DBLP:conf/emnlp/SocherPWCMNP13}, and the results were measured by accuracy of two-way classification.

The prompts for each task were illustrated in Table~\ref{tab-prompt}.

\begin{table*}[!htb]
\centering
\begin{tabular}{p{0.95\linewidth}}
\toprule

NLI:\\
\{\texttt{SENTENCE1}\} entails the \{\texttt{SENTENCE2}\}. True or False? answer:\\

\midrule

Open-domain QA:\\
Refer to the passage below and answer the following question. Passage: \{\texttt{DOCUMENT}\} Question: \{\texttt{QUESTION}\}\\

\midrule

Summarization:\\
\{\texttt{DIALOGUE}\} TL;DR:\\

\midrule

Sentiment analysis:\\
For each snippet of text, label the sentiment of the text as positive or negative. The answer should be exact 'positive' or 'negative'. text: \{\texttt{TEXT}\} answer:\\

\bottomrule
\end{tabular}
\caption{The prompts to LLMs for evaluating their zero-shot performance on these general tasks.}
\label{tab-prompt}
\end{table*}

\subsection{Hyper-parameters for Elastic Net}\label{hy}

In our experiment, we set \(\lambda = 5 \times 10^{-7} \), \(\mu = 5 \times 10^{-1} \) for GPT2-XL\cite{radford2019language} and \(\lambda = 5 \times 10^{-7} \), \(\mu = 1 \times 10^{-3} \) for LLaMA-3 (8B)\cite{llama3}.

\begin{figure*}[!hbt]
  \centering
  \includegraphics[width=0.5\textwidth]{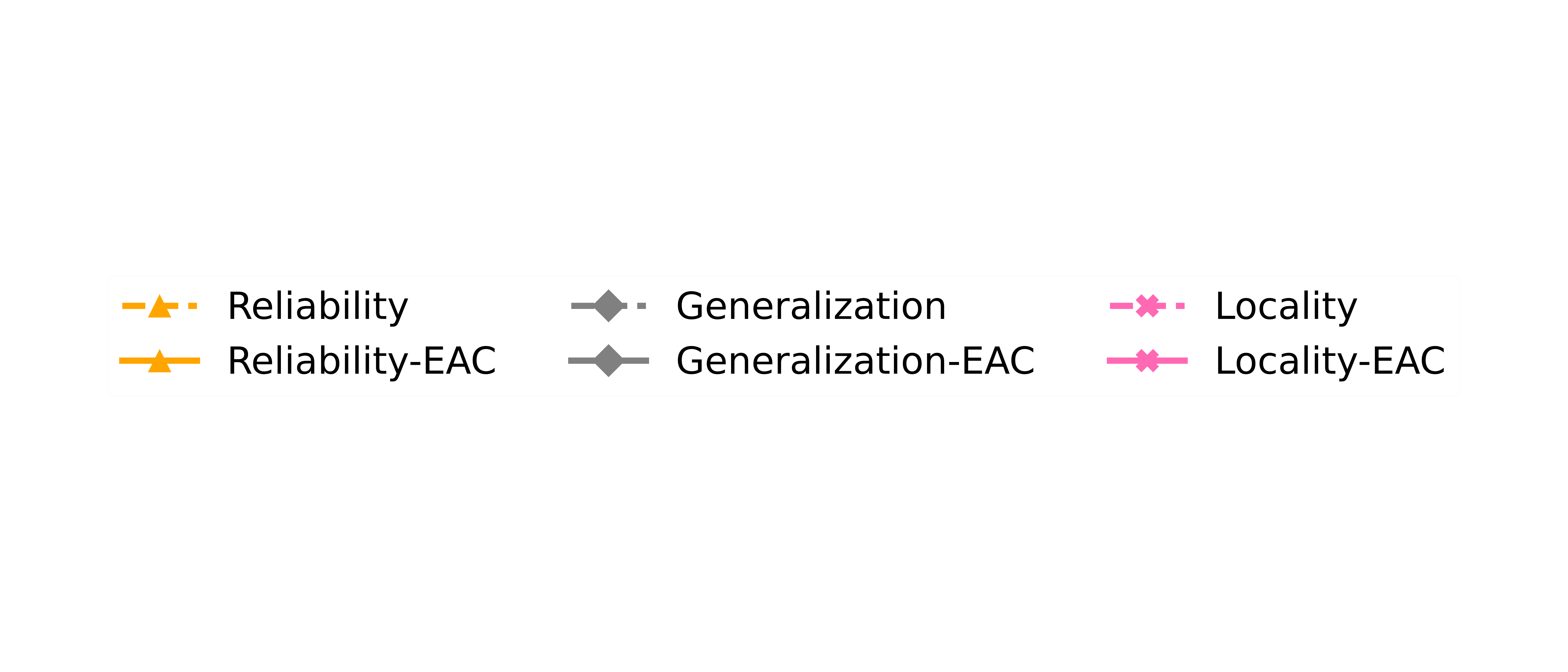}
  \vspace{-4mm}
\end{figure*}

\begin{figure*}[!hbt]
  \centering
  \subfigure{
  \includegraphics[width=0.23\textwidth]{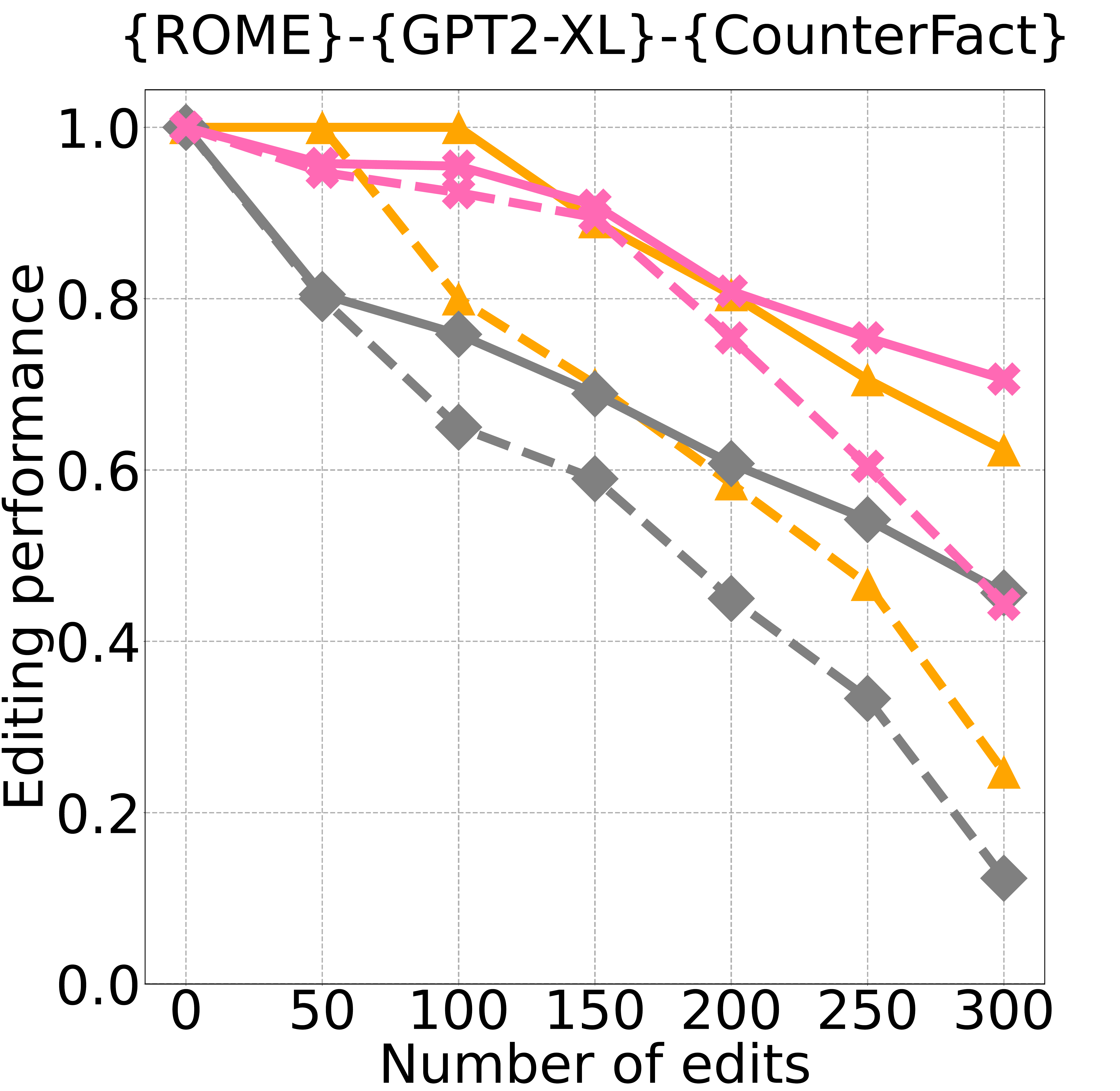}}
  \subfigure{
  \includegraphics[width=0.23\textwidth]{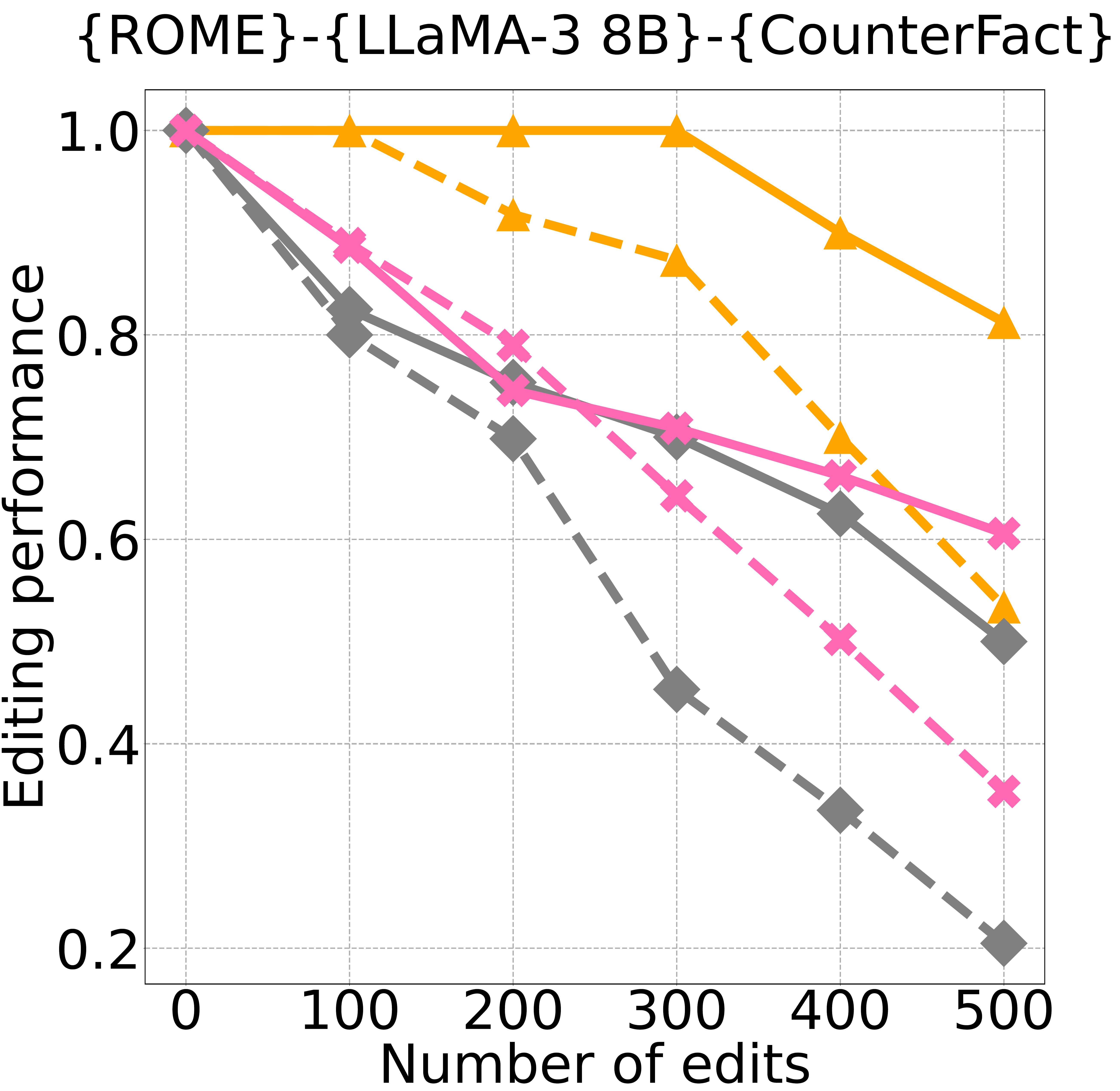}}
  \subfigure{
  \includegraphics[width=0.23\textwidth]{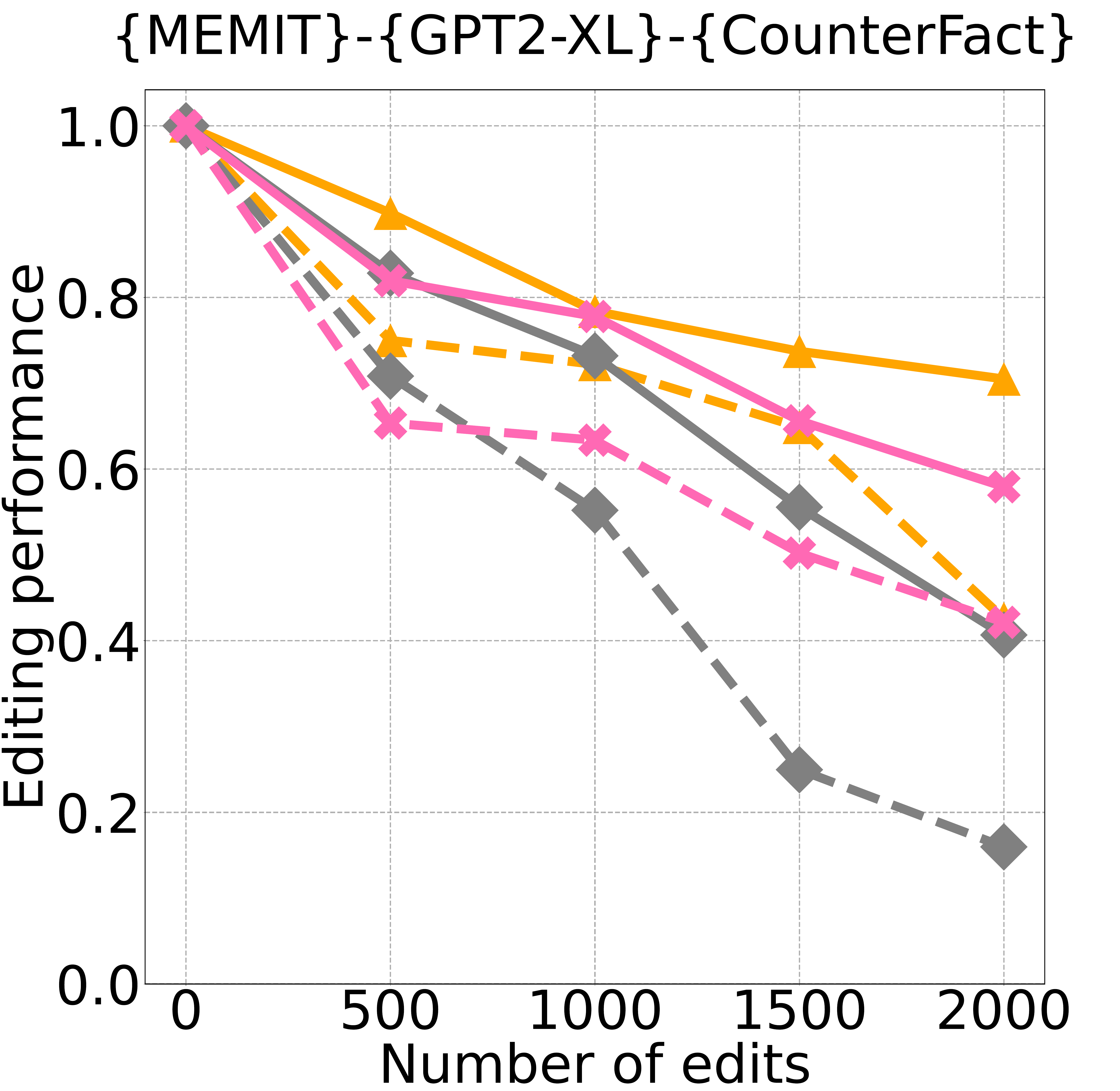}}
  \subfigure{
  \includegraphics[width=0.23\textwidth]{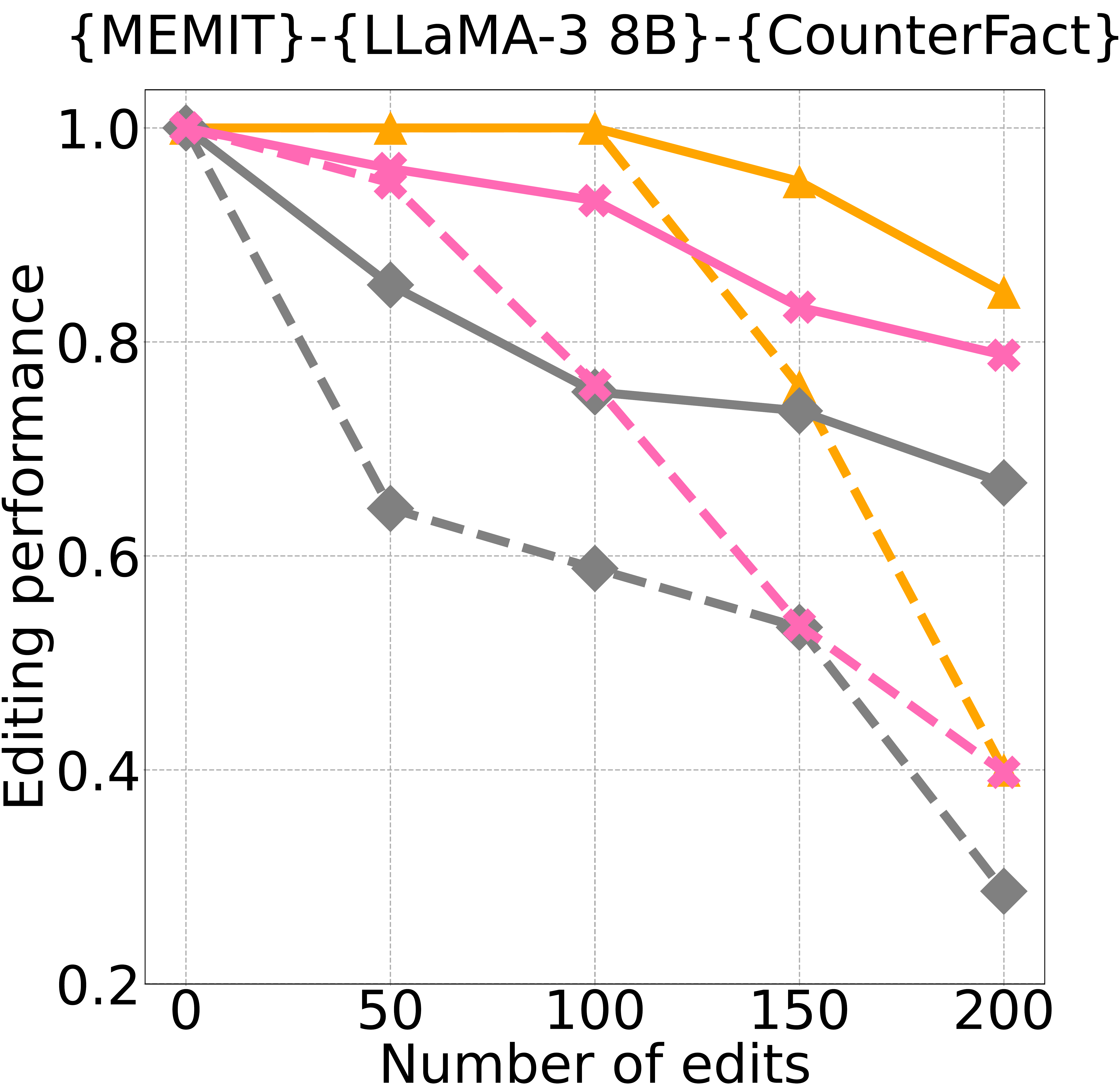}}
  \caption{Edited on CounterFact, editing performance of edited models using the ROME~\cite{DBLP:conf/nips/MengBAB22} and MEMIT~\cite{DBLP:conf/iclr/MengSABB23} on GPT2-XL~\cite{radford2019language} and LLaMA-3 (8B)~\cite{llama3}, as the number of edits increases before and after the introduction of EAC.}
  \vspace{-4mm}
  \label{edit-performance-cf}
\end{figure*}

\begin{figure*}[!hbt]
  \centering
  \includegraphics[width=0.75\textwidth]{figures/legend.pdf}
  \vspace{-4mm}
\end{figure*}

\begin{figure*}[!htb]
  \centering
  \subfigure{
  \includegraphics[width=0.23\textwidth]{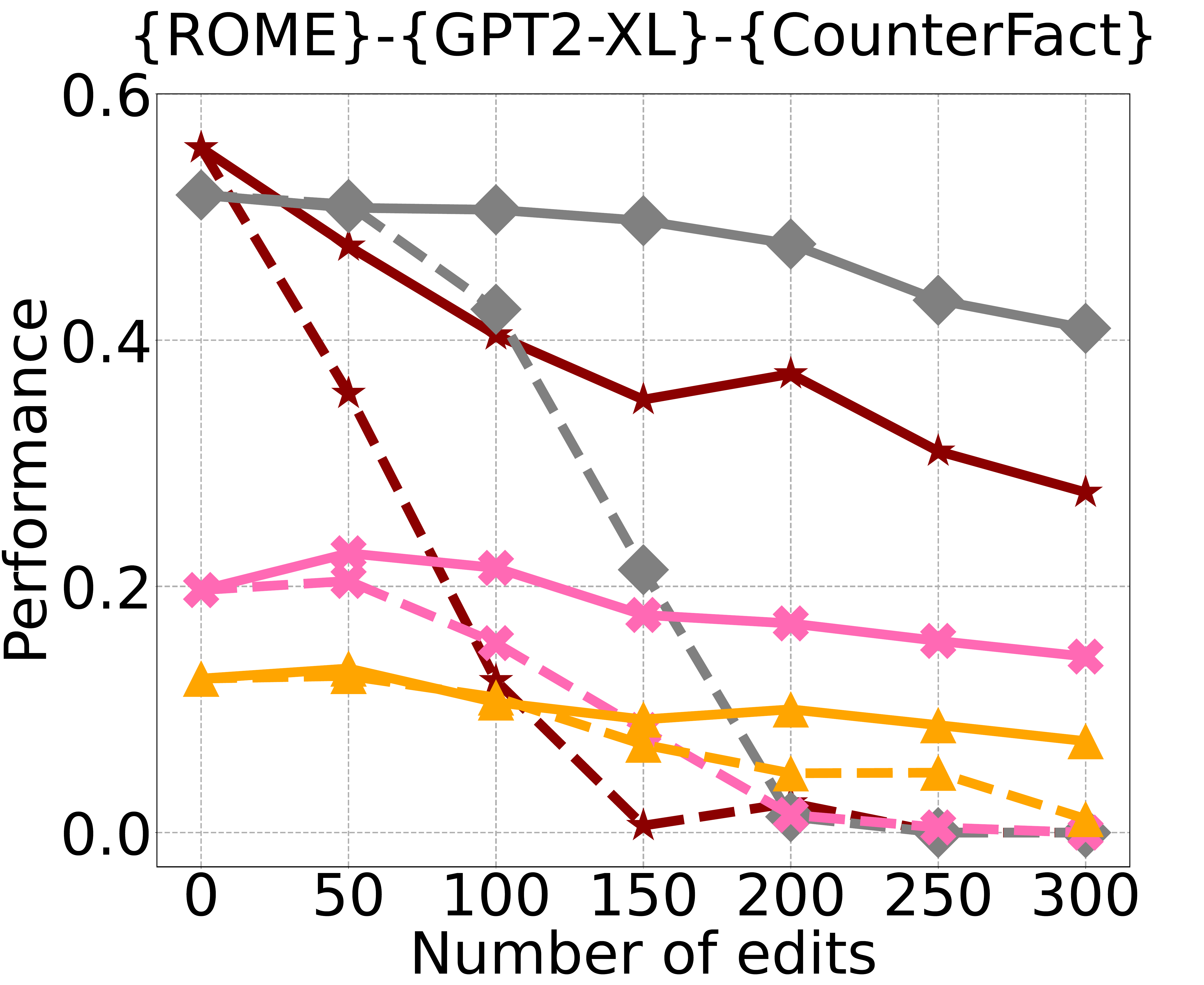}}
  \subfigure{
  \includegraphics[width=0.23\textwidth]{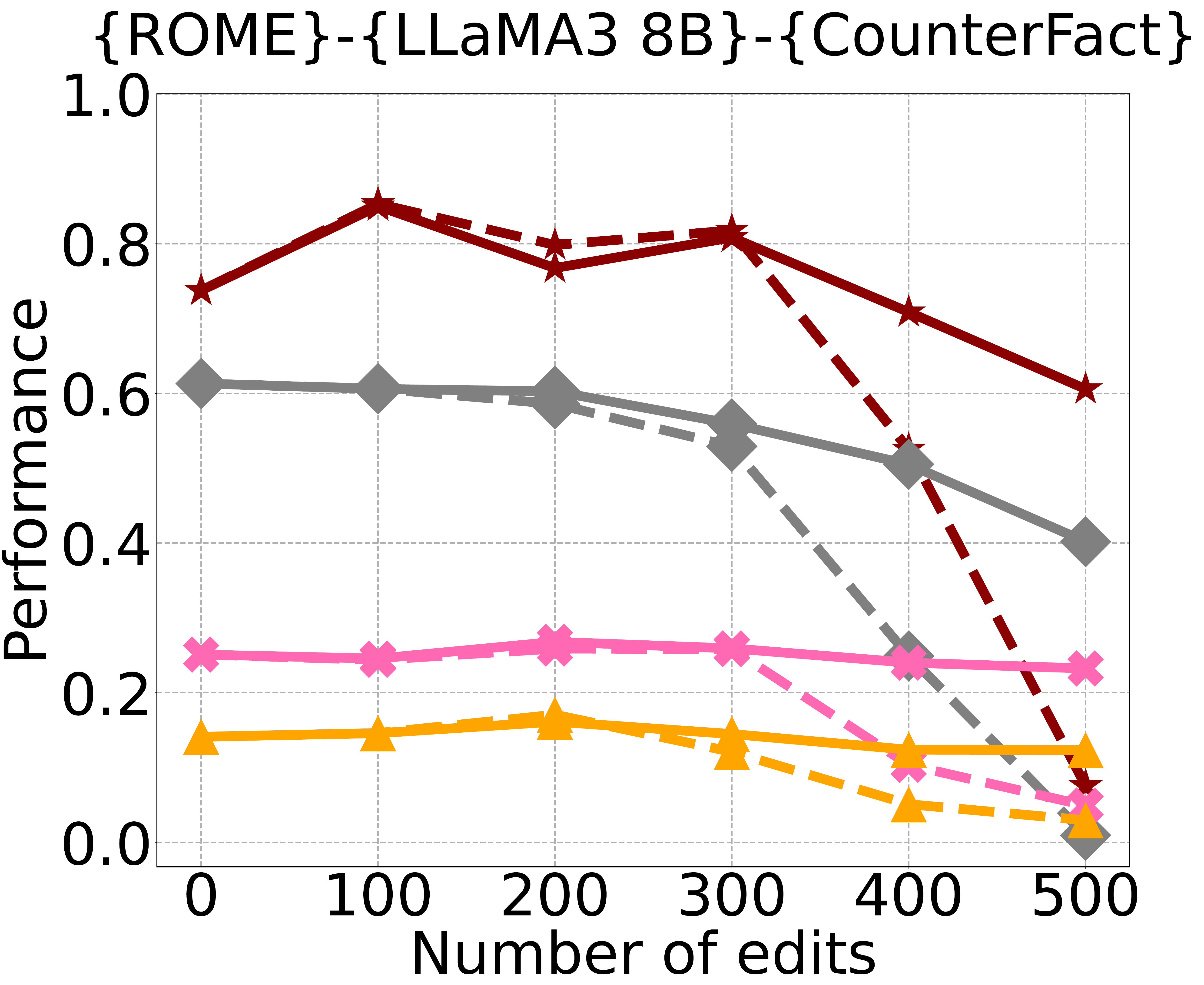}}
  \subfigure{
  \includegraphics[width=0.23\textwidth]{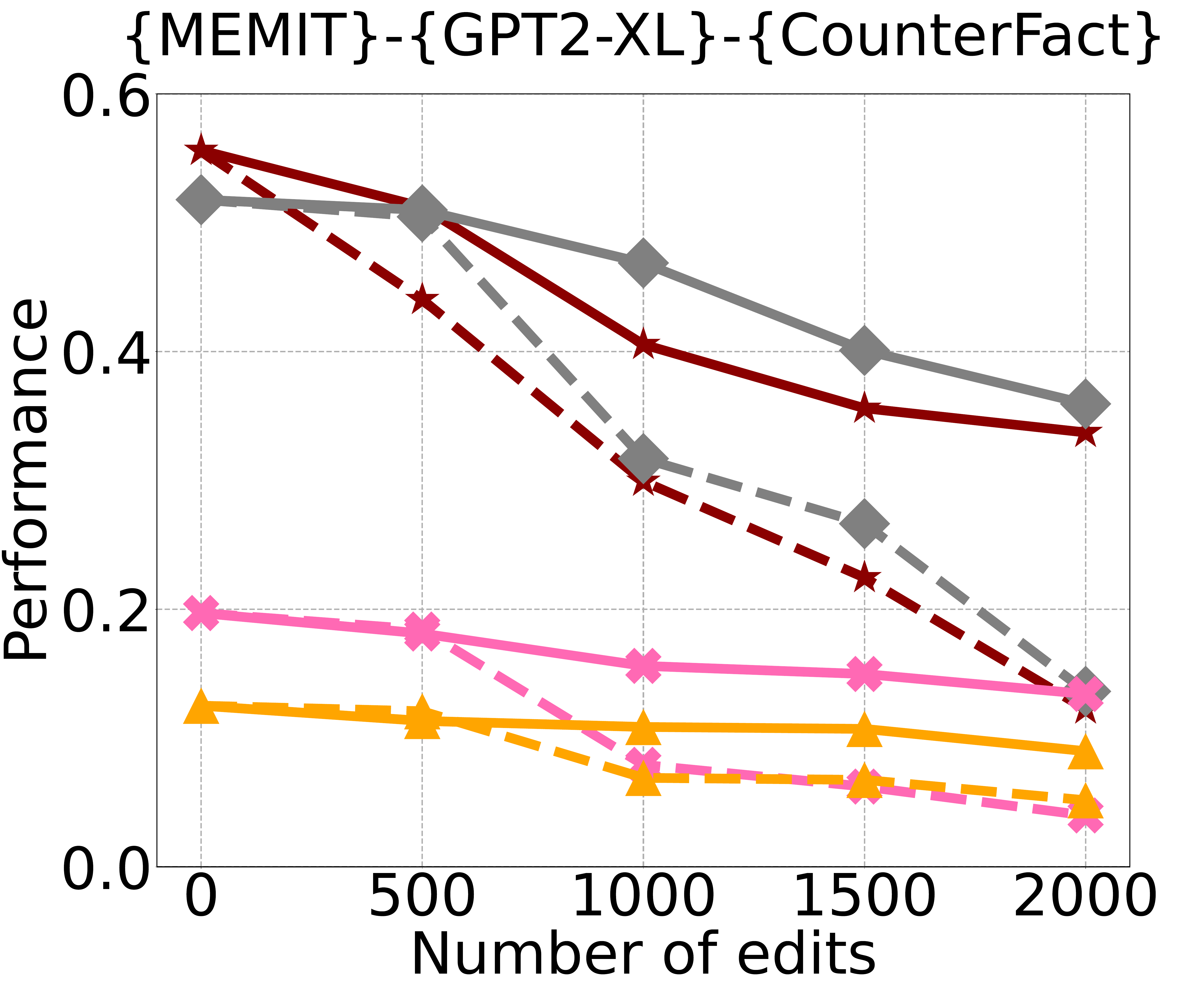}}
  \subfigure{
  \includegraphics[width=0.23\textwidth]{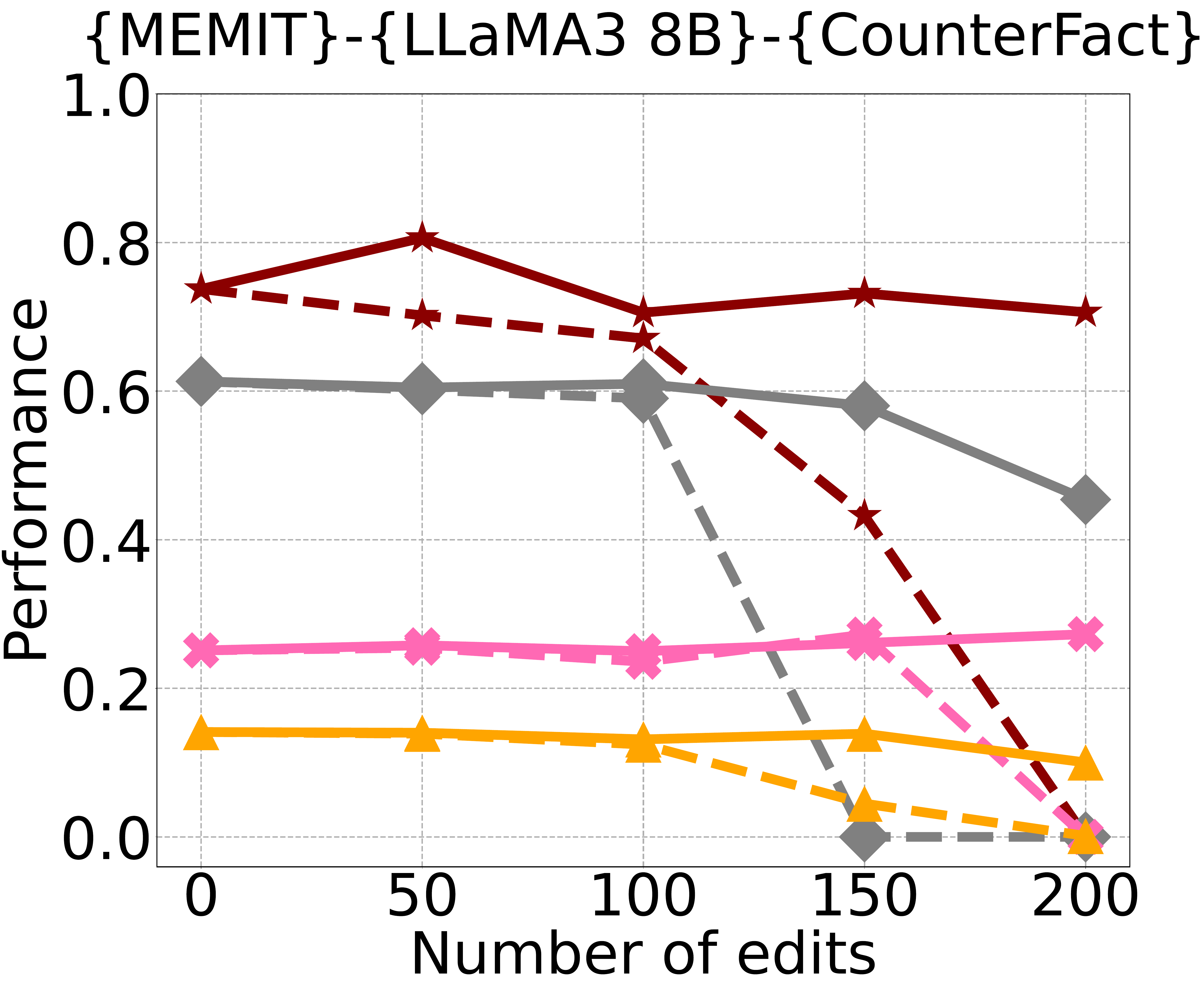}}
  \caption{Edited on CounterFact, performance on general tasks using the ROME~\cite{DBLP:conf/nips/MengBAB22} and MEMIT~\cite{DBLP:conf/iclr/MengSABB23} on GPT2-XL~\cite{radford2019language} and LLaMA-3 (8B)~\cite{llama3}, as the number of edits increases before and after the introduction of EAC.}
  \vspace{-4mm}
  \label{task-performance-cf}
\end{figure*}

\section{Experimental Results}\label{app}

\subsection{Results of Editing Performance}\label{cf-performance}
Applying CounterFact as the editing dataset, Figure~\ref{edit-performance-cf} presents the editing performance of the ROME~\cite{DBLP:conf/nips/MengBAB22} and MEMIT~\cite{DBLP:conf/iclr/MengSABB23} methods on GPT2-XL~\cite{radford2019language} and LLaMA-3 (8B)~\cite{llama3}, respectively, as the number of edits increases before and after the introduction of EAC.
The dashed line represents the ROME or MEMIT, while the solid line represents the ROME or MEMIT applying the EAC.

\subsection{Results of General Abilities}\label{cf-ability}
Applying CounterFact as the editing dataset, Figure~\ref{task-performance-cf} presents the performance on general tasks of edited models using the ROME~\cite{DBLP:conf/nips/MengBAB22} and MEMIT~\cite{DBLP:conf/iclr/MengSABB23} methods on GPT2-XL~\cite{radford2019language} and LLaMA-3 (8B)~\cite{llama3}, respectively, as the number of edits increases before and after the introduction of EAC. 
The dashed line represents the ROME or MEMIT, while the solid line represents the ROME or MEMIT applying the EAC.

\subsection{Results of Larger Model}\label{13 B}
\begin{figure*}[!hbt]
  \centering
  \subfigure{
  \includegraphics[width=0.23\textwidth]{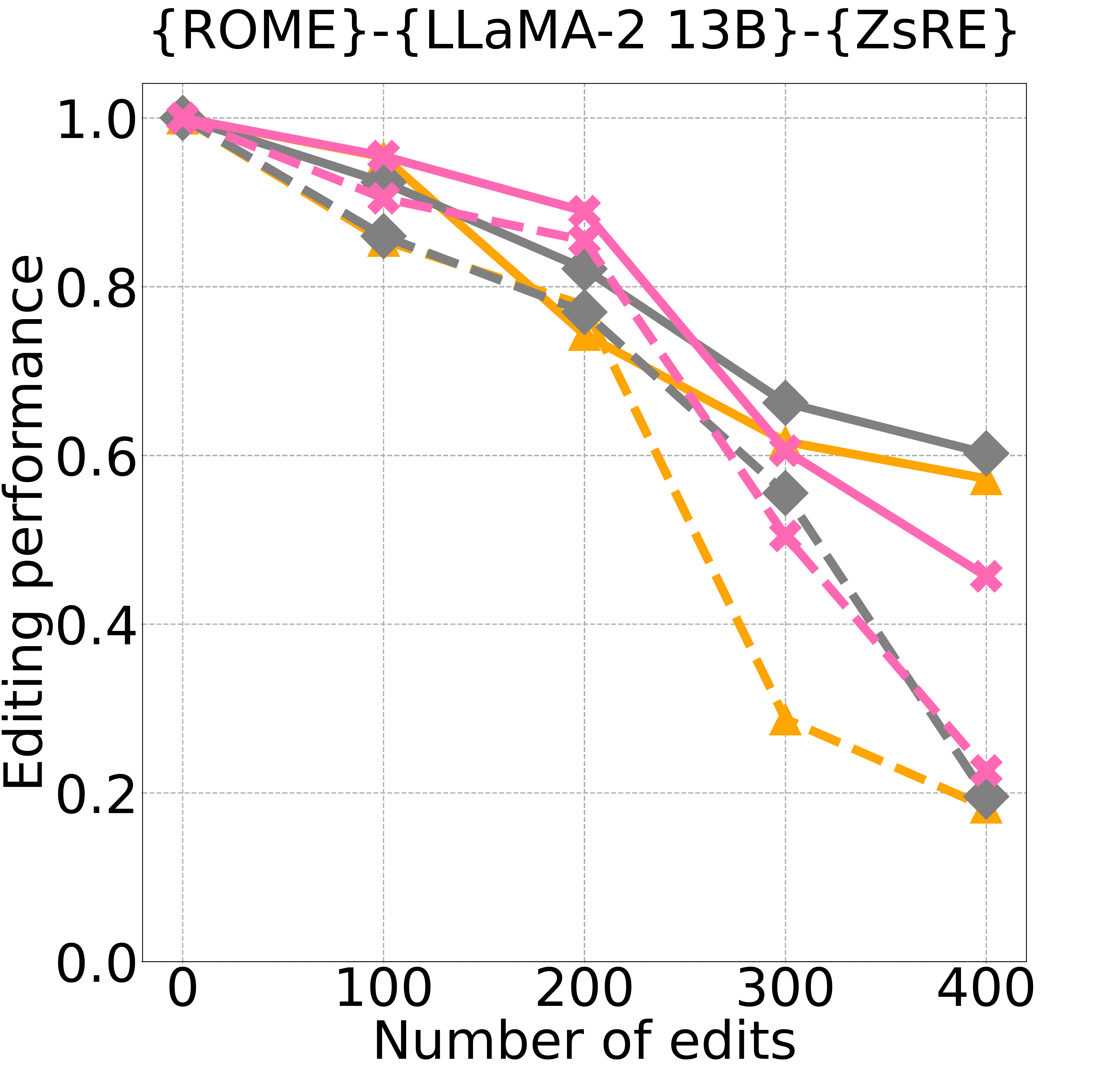}}
  \subfigure{
  \includegraphics[width=0.23\textwidth]{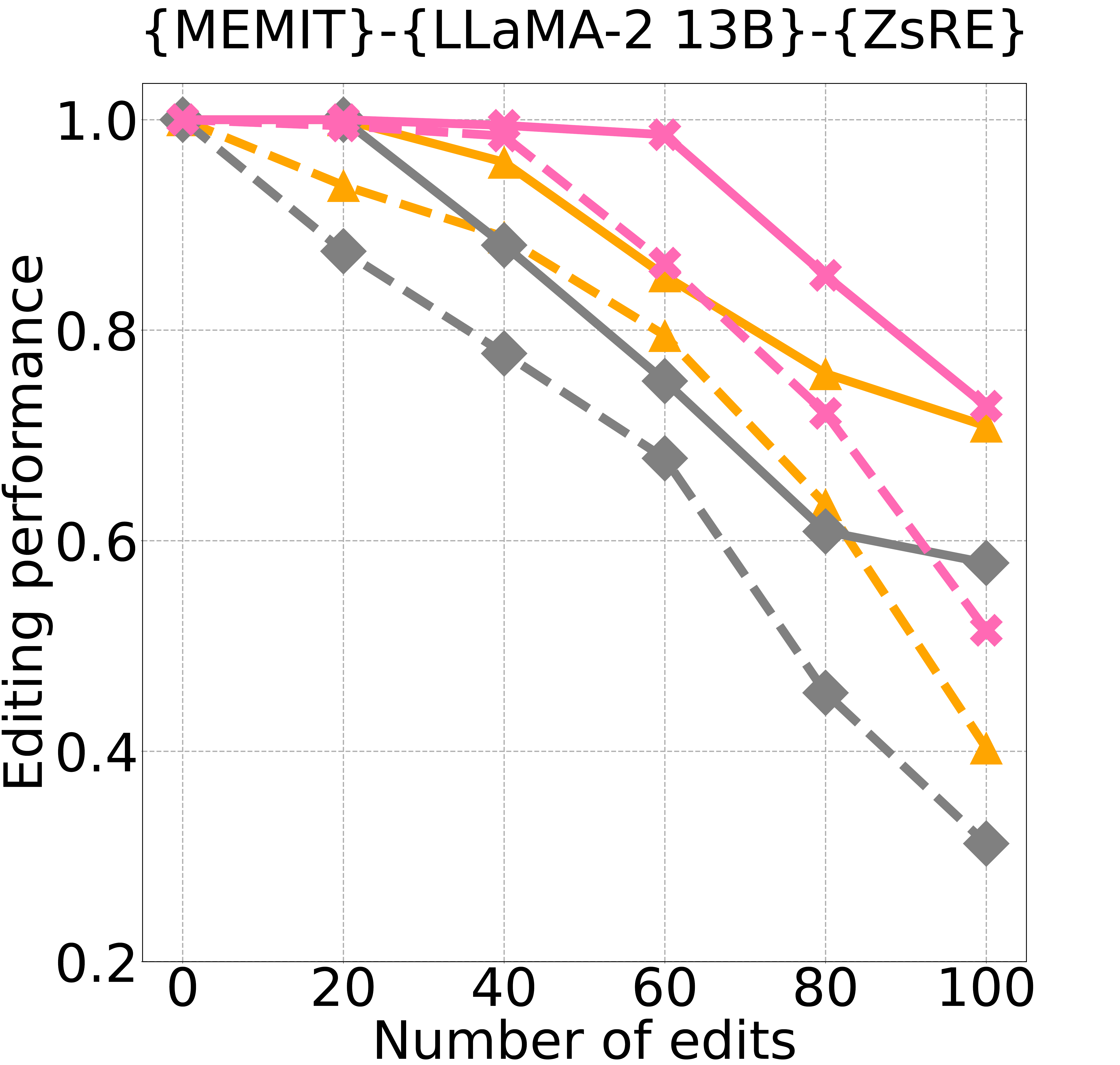}}
  \subfigure{
  \includegraphics[width=0.23\textwidth]{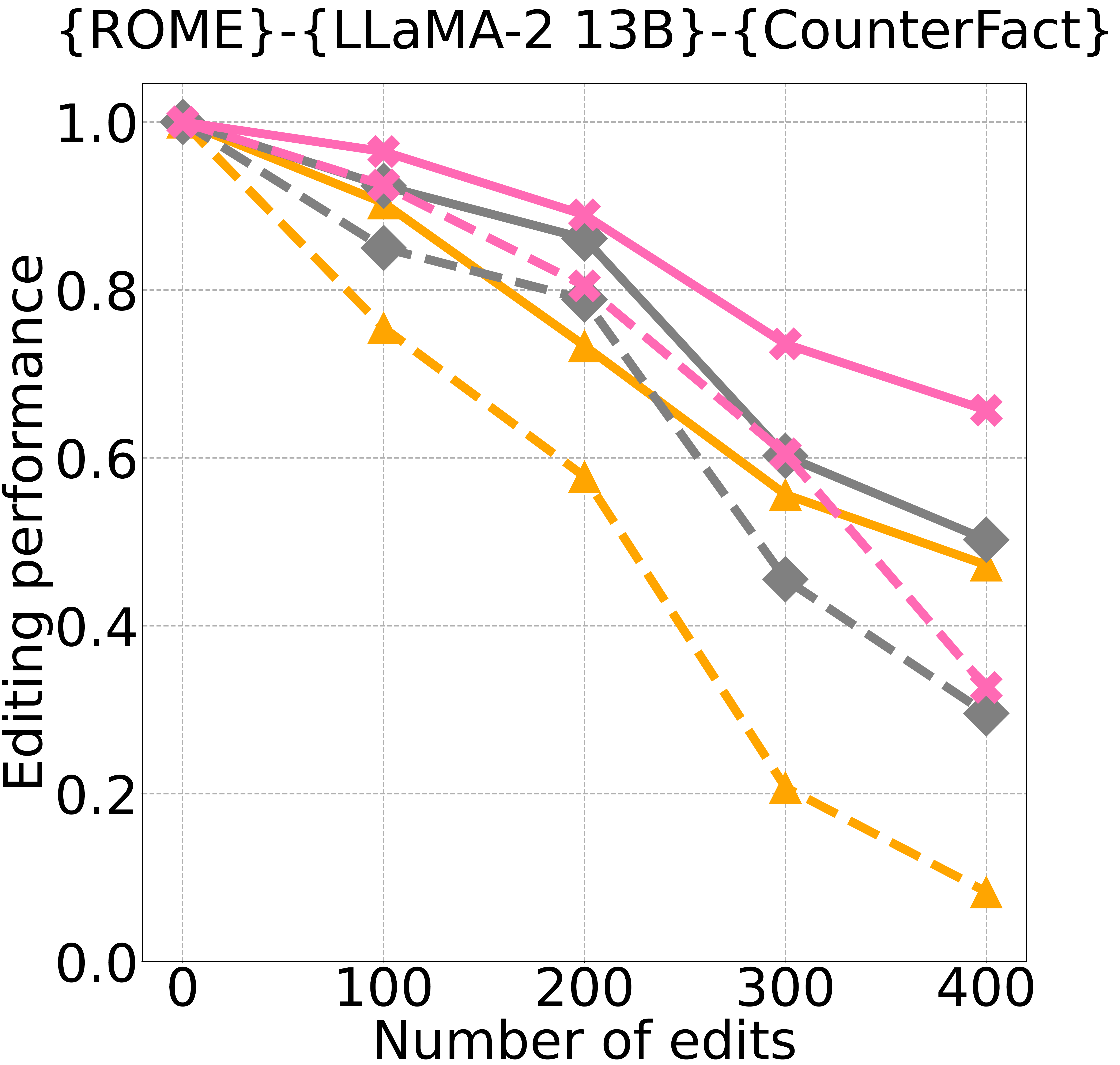}}
  \subfigure{
  \includegraphics[width=0.23\textwidth]{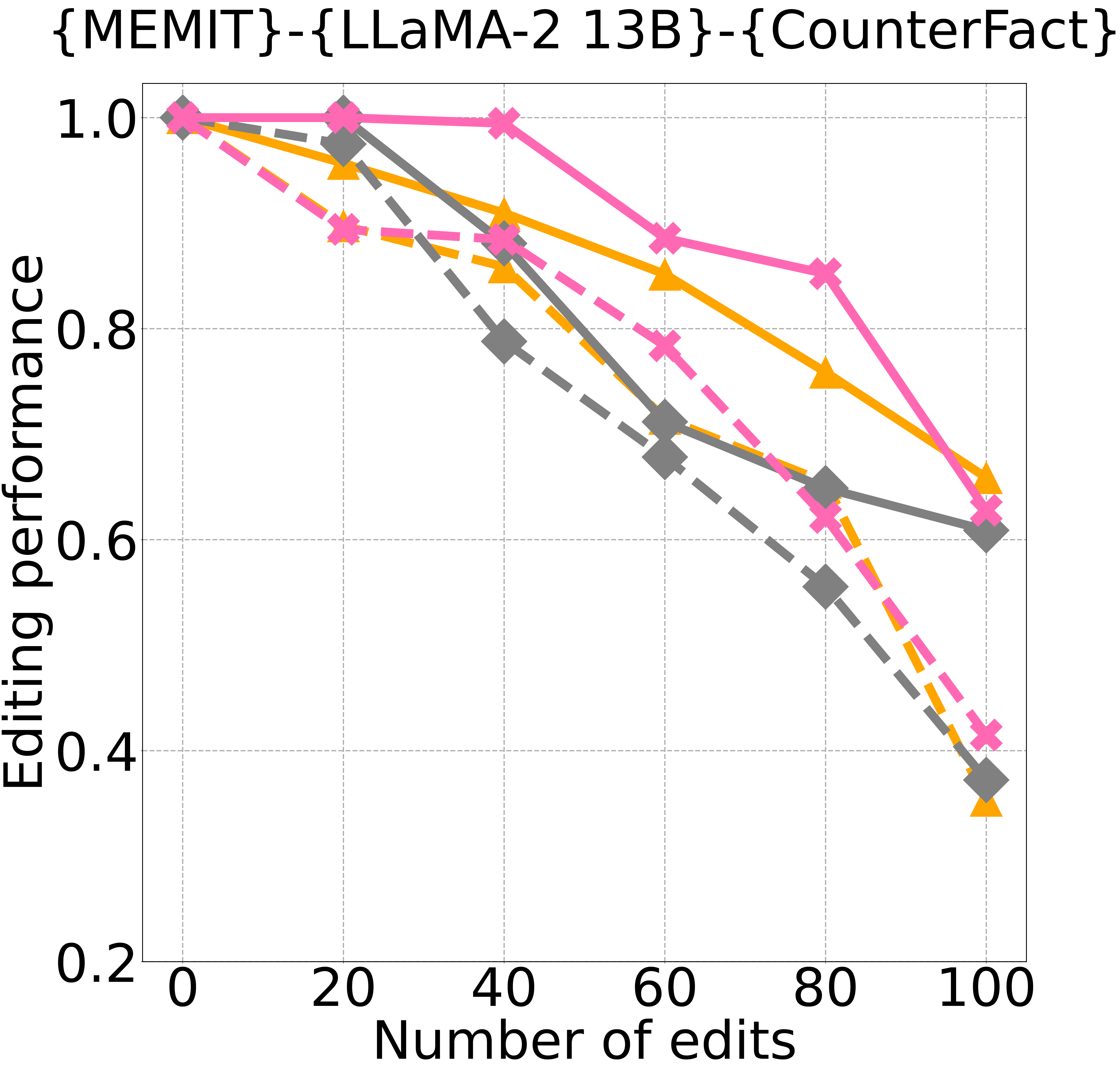}}
  \caption{Editing performance of edited models using the ROME~\cite{DBLP:conf/nips/MengBAB22} and MEMIT~\cite{DBLP:conf/iclr/MengSABB23} on LLaMA-2 (13B)~\cite{DBLP:journals/corr/abs-2307-09288}, as the number of edits increases before and after the introduction of EAC.}
  \vspace{-4mm}
  \label{13B-edit}
\end{figure*}

\begin{figure*}[!htb]
  \centering
  \subfigure{
  \includegraphics[width=0.23\textwidth]{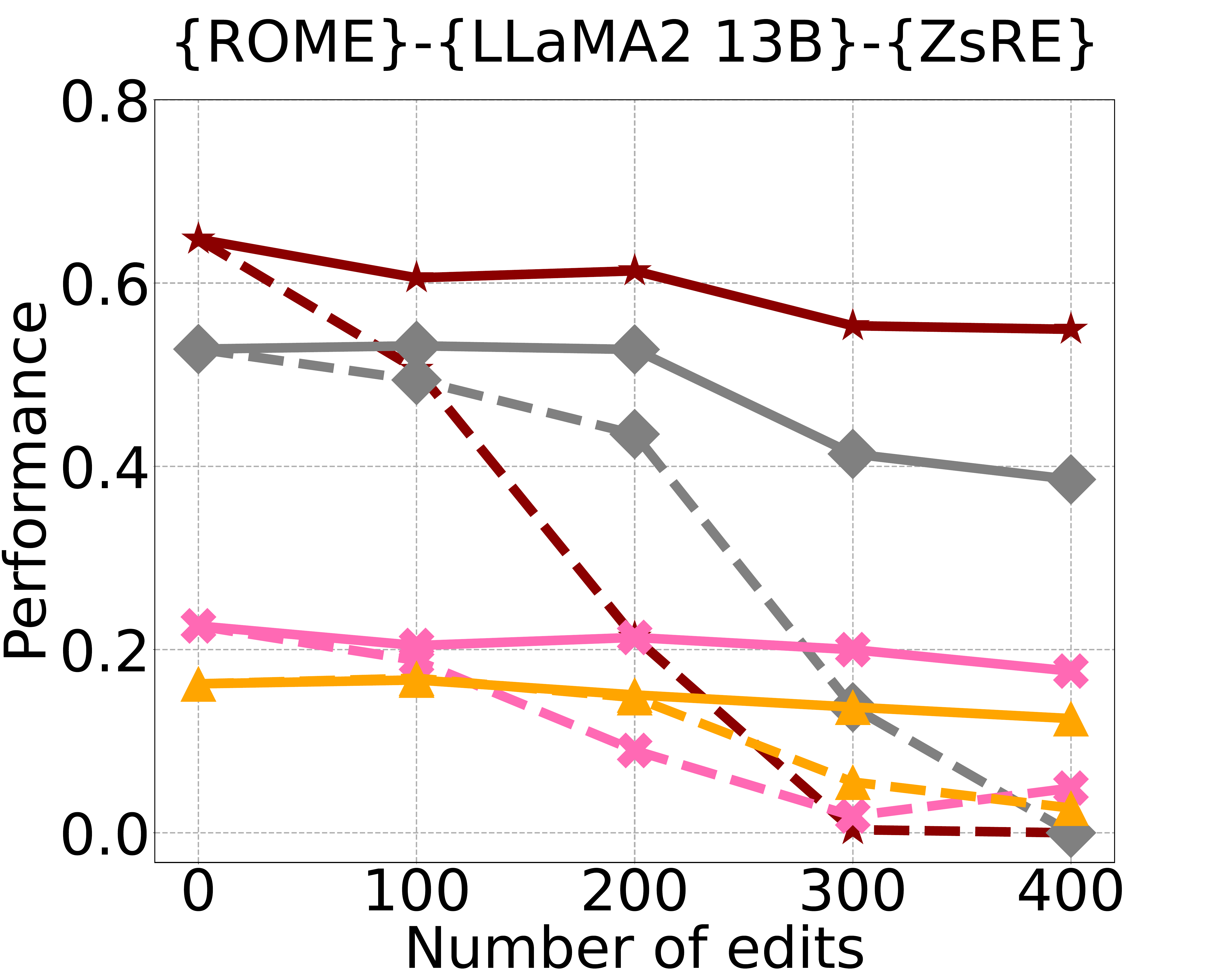}}
  \subfigure{
  \includegraphics[width=0.23\textwidth]{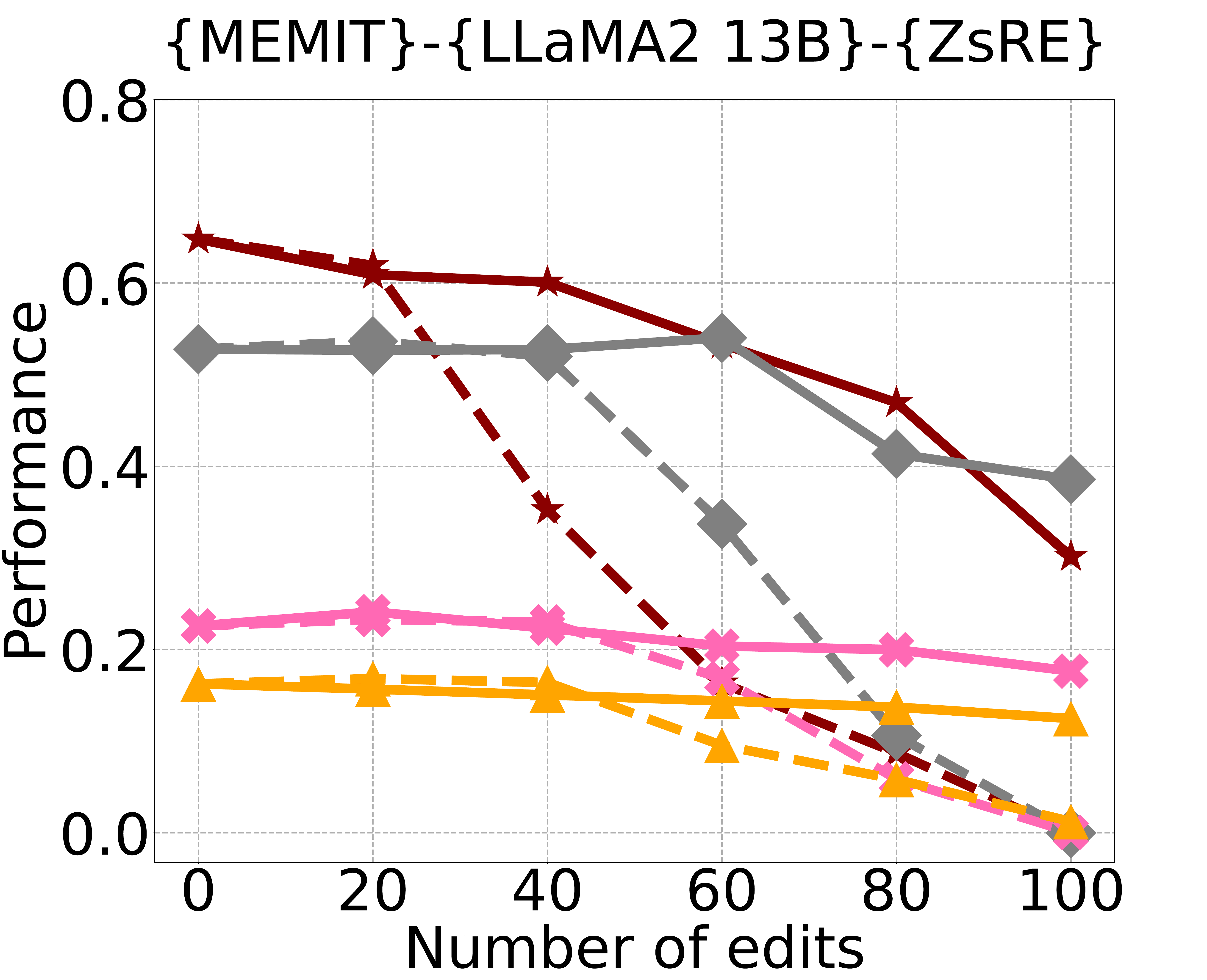}}
  \subfigure{
  \includegraphics[width=0.23\textwidth]{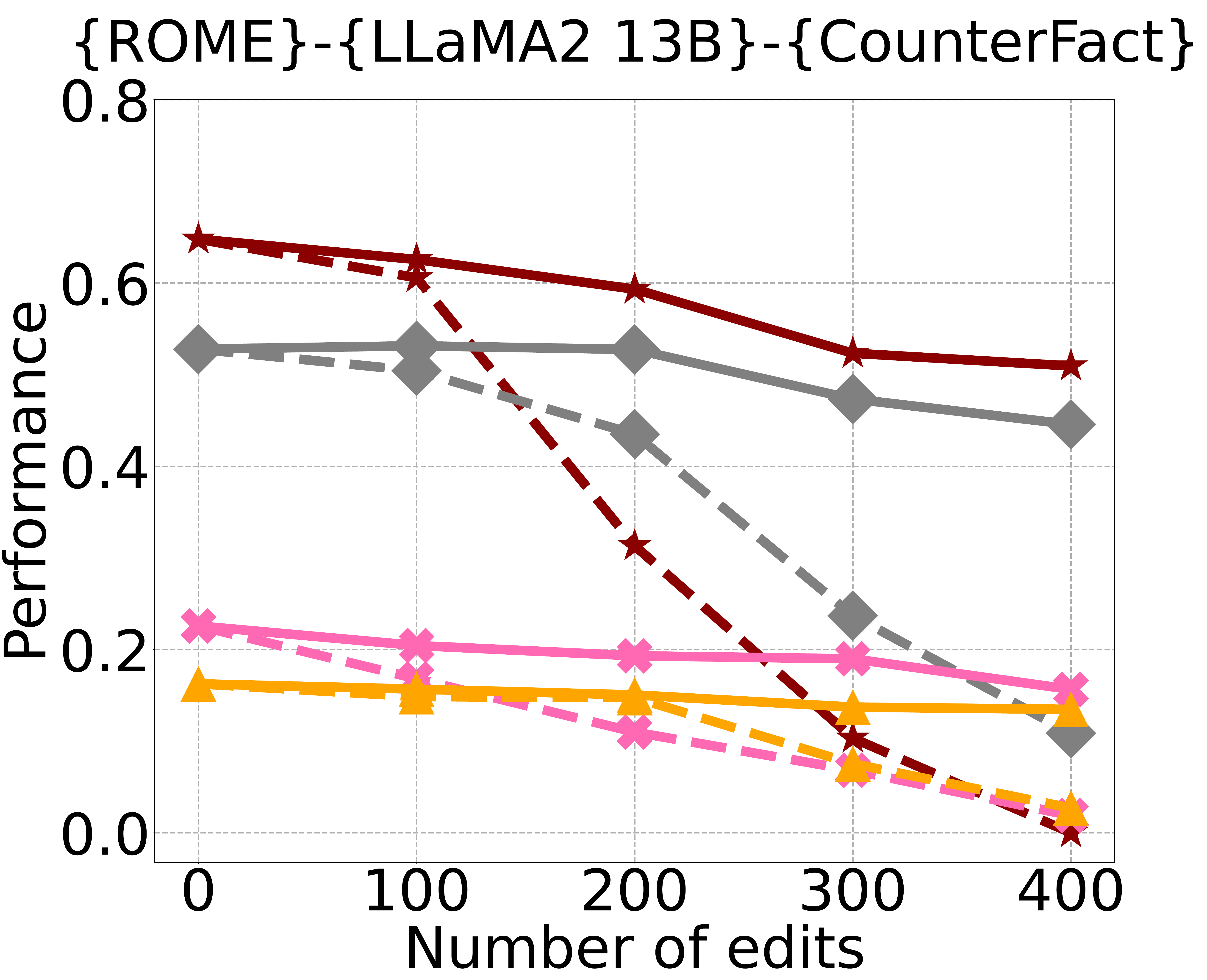}}
  \subfigure{
  \includegraphics[width=0.23\textwidth]{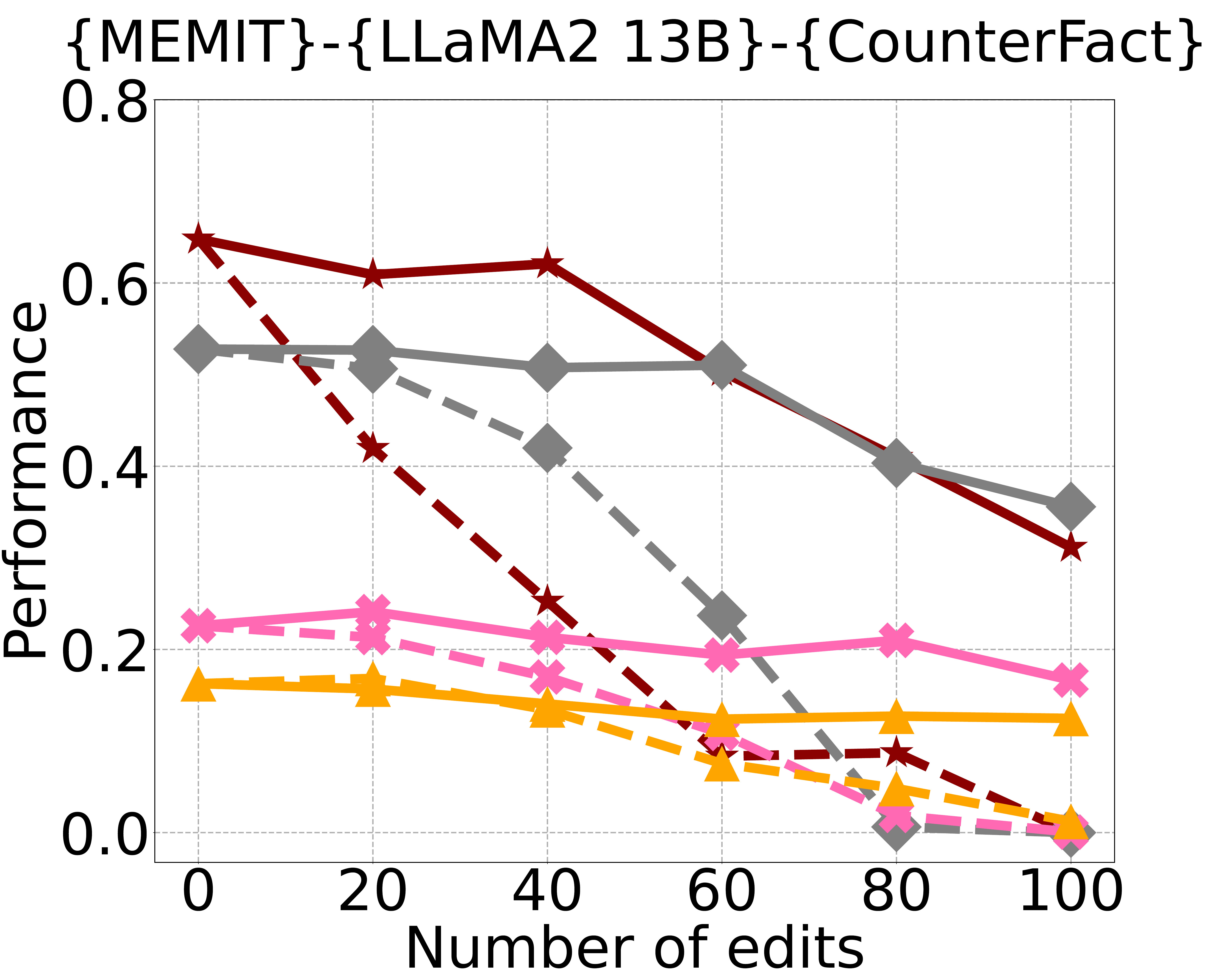}}
  \caption{Performance on general tasks using the ROME~\cite{DBLP:conf/nips/MengBAB22} and MEMIT~\cite{DBLP:conf/iclr/MengSABB23} on LLaMA-2 (13B)~\cite{DBLP:journals/corr/abs-2307-09288}, as the number of edits increases before and after the introduction of EAC.}
  \vspace{-4mm}
  \label{13B}
\end{figure*}

To better demonstrate the scalability and efficiency of our approach, we conducted experiments using the LLaMA-2 (13B)~\cite{DBLP:journals/corr/abs-2307-09288}.
Figure~\ref{13B-edit} presents the editing performance of the ROME~\cite{DBLP:conf/nips/MengBAB22} and MEMIT~\cite{DBLP:conf/iclr/MengSABB23} methods on LLaMA-2 (13B)
~\cite{DBLP:journals/corr/abs-2307-09288}, as the number of edits increases before and after the introduction of EAC.
Figure~\ref{13B} presents the performance on general tasks of edited models using the ROME and MEMIT methods on LLaMA-2 (13B), as the number of edits increases before and after the introduction of EAC.
The dashed line represents the ROME or MEMIT, while the solid line represents the ROME or MEMIT applying the EAC.

\begin{figure*}[!hbt]
  \centering
  \includegraphics[width=0.5\textwidth]{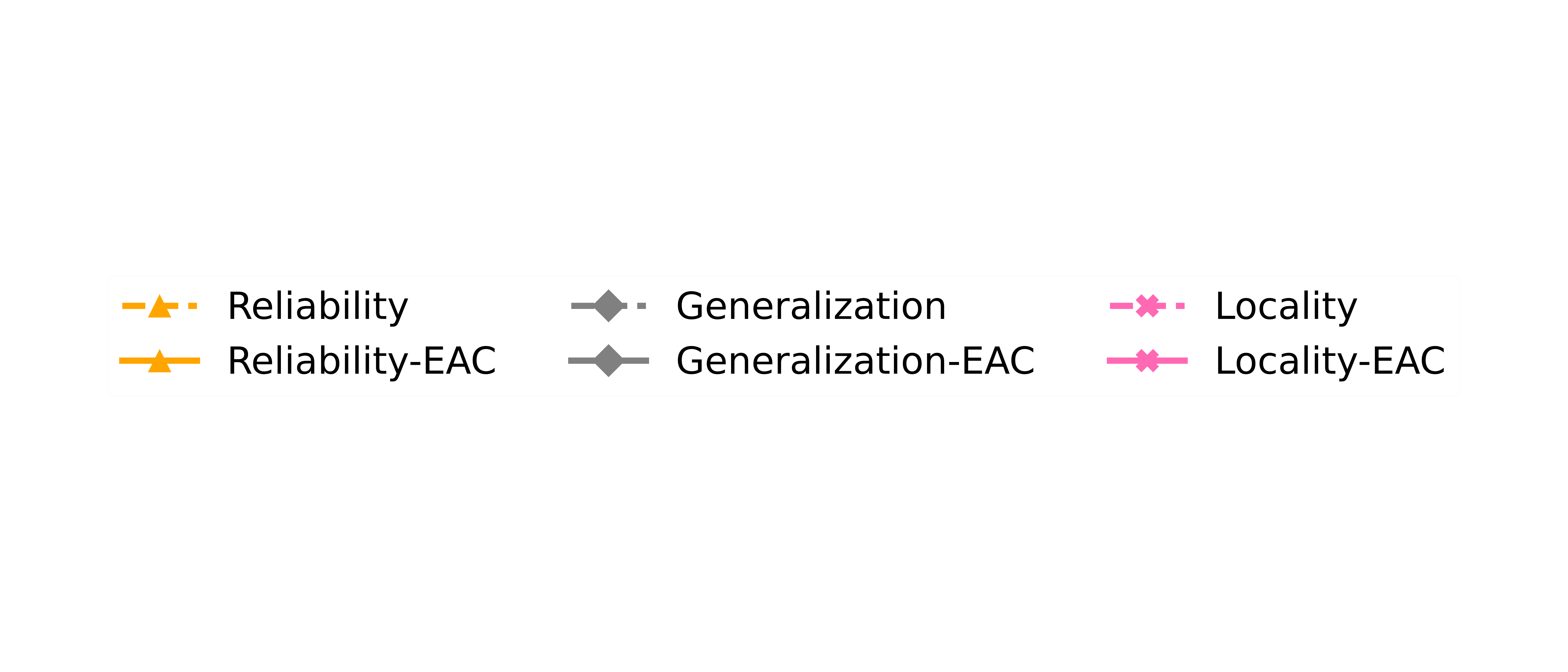}
  \vspace{-4mm}
\end{figure*}

\begin{figure*}[!hbt]
  \centering
  \includegraphics[width=0.75\textwidth]{figures/legend.pdf}
  \vspace{-4mm}
\end{figure*}

\section{Analysis of Elastic Net}
\label{FT}
\begin{figure*}[t]
  \centering
  \subfigure{
  \includegraphics[width=0.22\textwidth]{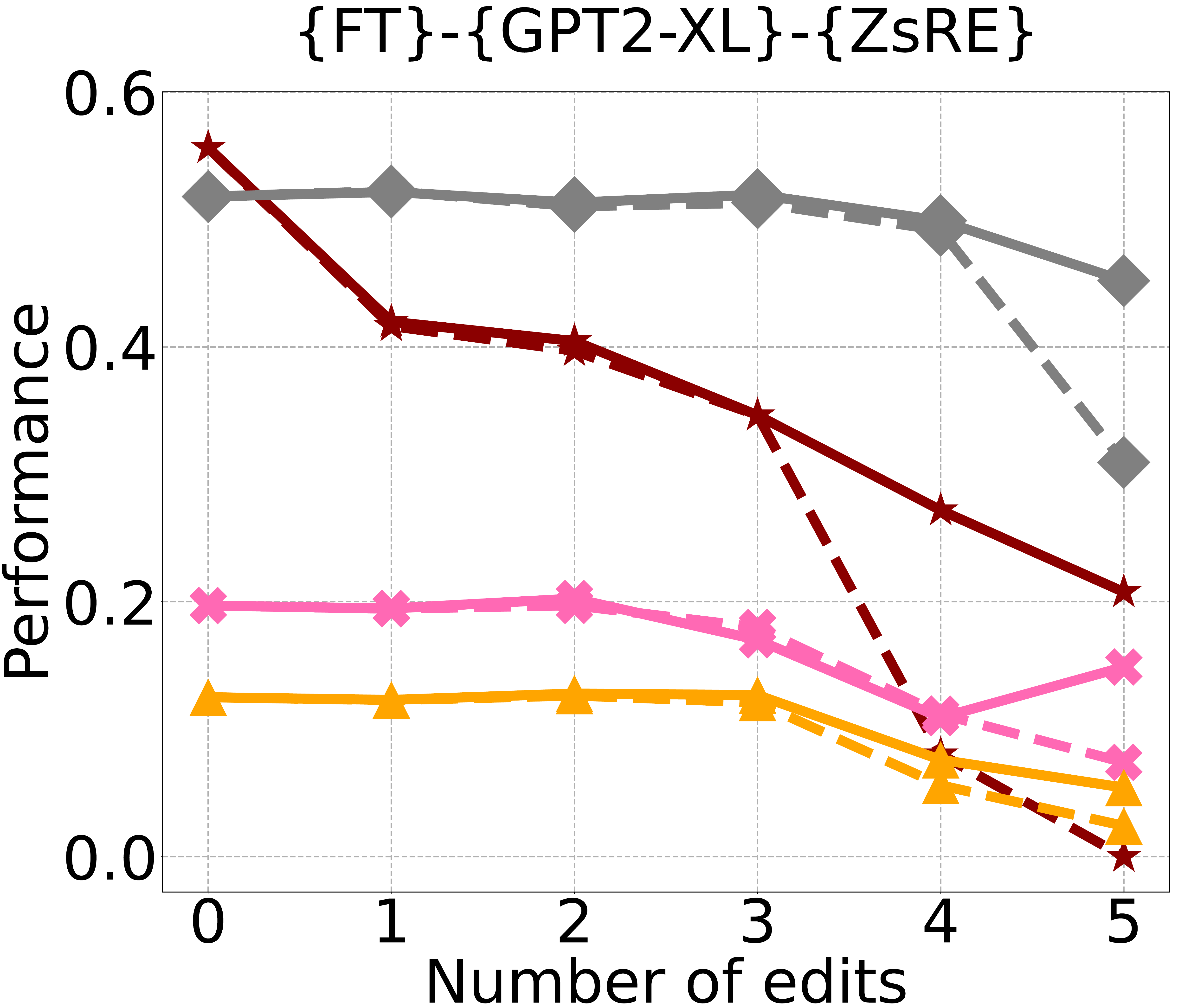}}
  \subfigure{
  \includegraphics[width=0.22\textwidth]{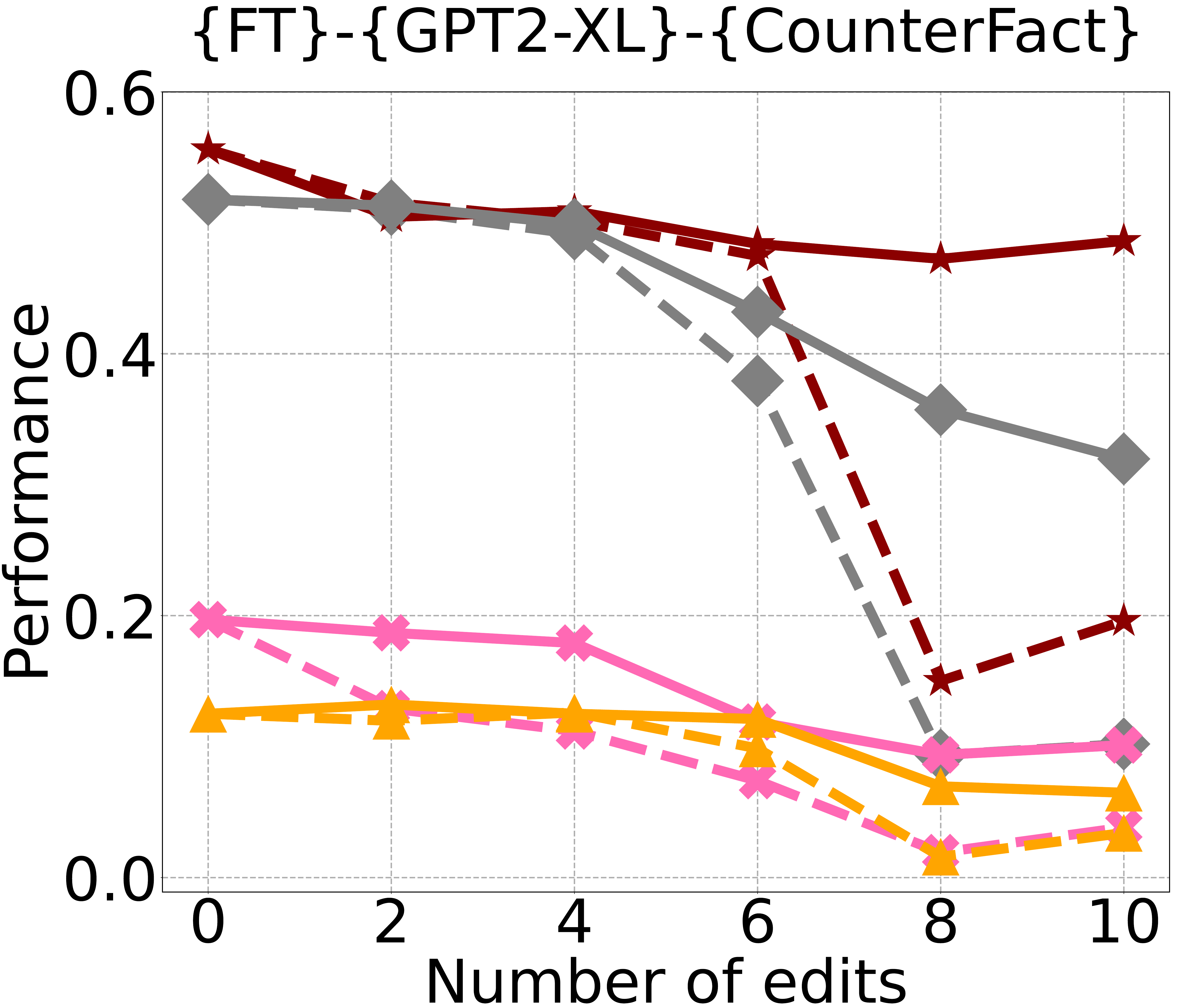}}
  \vspace{-2mm}
  \caption{Edited on the ZsRE or CounterFact datasets, the sequential editing performance of FT~\cite{DBLP:conf/emnlp/CaoAT21} and FT with elastic net on GPT2-XL before and after the introduction of elastic net.}
  \vspace{-2mm}
  \label{ft}
\end{figure*}

It is worth noting that the elastic net introduced in EAC can be applied to methods beyond ROME and MEMIT, such as FT~\cite{DBLP:conf/emnlp/CaoAT21}, to preserve the general abilities of the model.
Unlike the previously mentioned fine-tuning, FT is a model editing approach. It utilized the gradient to gather information about the knowledge to be updated and applied this information directly to the model parameters for updates.
Similar to the approaches of ROME and MEMIT, which involve locating parameters and modifying them, the FT method utilizes gradient information to directly update the model parameters for editing. Therefore, we incorporate an elastic net during the training process to constrain the deviation of the edited matrix.
Figure~\ref{ft} shows the sequential editing performance of FT on GPT2-XL and LLaMA-3 (8B) before and after the introduction of elastic net.
The dashed line represents the FT, while the solid line represents the FT applying the elastic net.
The experimental results indicate that when using the FT method to edit the model, the direct use of gradient information to modify the parameters destroys the general ability of the model. By constraining the deviation of the edited matrix, the incorporation of the elastic net effectively preserves the general abilities of the model.

\begin{figure*}[t]
  \centering
  \subfigure{
  \includegraphics[width=0.43\textwidth]{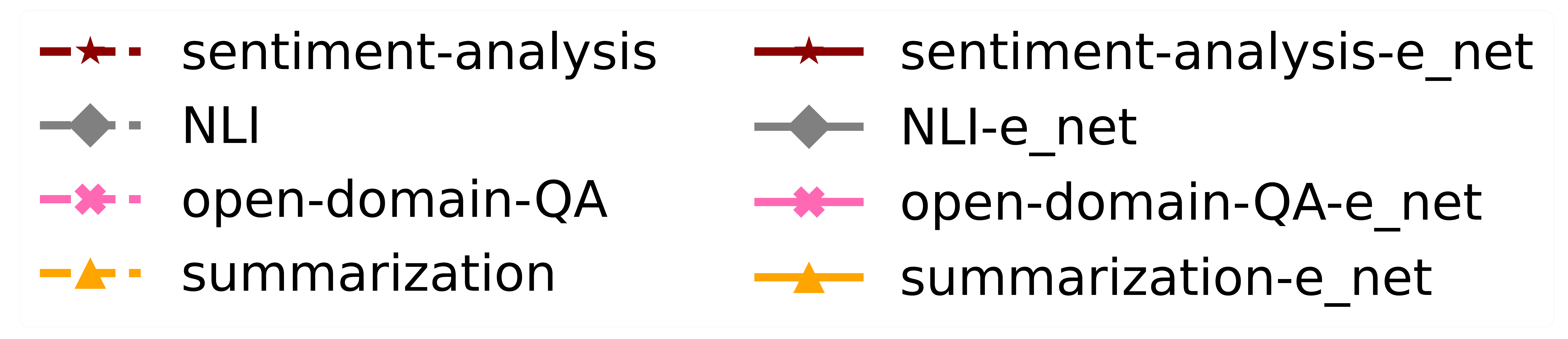}}
\end{figure*}

\end{document}